\documentclass[3p]{elsarticle}
\usepackage{lineno,hyperref}
\modulolinenumbers[5]

\journal{Journal Knowledge base systems}

\usepackage{ntheorem}

\newtheorem{obs}{Observation}

\usepackage{amsmath}
\usepackage{subcaption}

\graphicspath{{figs/pdf/}{figs/jpeg/}{figs/png/}{figs/svg/}{figs/bio/}}
\DeclareGraphicsExtensions{.pdf,.jpeg,.png,.svg}

\usepackage[export]{adjustbox}

\hypersetup{
	colorlinks,
	linkcolor={red!50!black},
	citecolor={blue!50!black},
	urlcolor={blue!80!black}
}

\usepackage[justification=centering]{caption}

\usepackage{array,etoolbox}
\preto\tabular{\setcounter{magicrownumbers}{0}}
\newcounter{magicrownumbers}
\def\rownumber{}

\usepackage{soul}

\input{insbox}

\begin{document}

\begin{frontmatter}

\title{Exploiting Non-Taxonomic Relations for Measuring Semantic Similarity and Relatedness in WordNet}

\author[firstaddress]{Mohannad AlMousa\corref{mycorrespondingauthor}}
\ead{malmous@lakeheadu.ca}
\cortext[mycorrespondingauthor]{Corresponding author.}

\author[firstaddress]{Rachid Benlamri}
\ead{rbenlamr@lakeheadu.ca}

\author[secondaddress]{Richard Khoury}
\ead{richard.khoury@ift.ulaval.ca}

\address[firstaddress]{Department of Software Engineering, Lakehead University, Thunder Bay,ON, P7B 5E1, Canada.}
\address[secondaddress]{Department of Computer Science and Software Engineering, Université Laval, Québec, QC G1V 0A6, Canada.}

\begin{abstract}
Various applications in the areas of computational linguistics and artificial intelligence employ semantic similarity to solve challenging tasks, such as word sense disambiguation, text classification, information retrieval, machine translation, and document clustering. Previous work on semantic similarity followed a mono-relational approach using mostly the taxonomic relation "ISA". This paper explores the benefits of using all types of non-taxonomic relations in large linked data, such as WordNet knowledge graph, to enhance existing semantic similarity and relatedness measures. We propose a holistic poly-relational approach based on a new relation-based information content and non-taxonomic-based weighted paths to devise a comprehensive semantic similarity and relatedness measure. To demonstrate the benefits of exploiting non-taxonomic relations in a knowledge graph, we used three strategies to deploy non-taxonomic relations at different granularity levels. We conducted experiments on four well-known gold standard datasets, and the results demonstrated the robustness and scalability of the proposed semantic similarity and relatedness measure, which significantly improves existing similarity measures. 
\end{abstract}

\begin{keyword}
\texttt{Semantic Similarity and Relatedness}\sep \texttt{Knowledge graph}\sep \texttt{Information content}\sep \texttt{WordNet}.
\end{keyword}

\end{frontmatter}

\section{Introduction}\label{sec:introduction}
The rapid expansion of Open Linked Data (OLD) requires a comprehensive semantic similarity measure that is yet to exist. DBpedia \cite{dbpedia}, Freebase \cite{freebase}, YAGO \cite{mahdisoltani2013yago3}, and WordNet \cite{miller1995wordnet} are examples of Knowledge Graphs (KG) that resemble OLD. KGs are representative of an ontological schema, which semantically models a specific domain of knowledge. 
In technical terms, ontology is a formal semantic representation of the concepts within a specific domain. The semantic representation is established through a set of axioms. An axiom connects two concepts and/or instances through a specific relation that models real-world connection in the form of \textit{subject}, \textit{predicate}, and \textit{object}. An interconnected set of axioms forms a knowledge base (KB), knowledge graph (KG), or semantic graph (SG) as referred to in \cite{ehrlinger2016_KGDef, barthelemy2005, eliassi2005}. Ehrlinger and W{\"o}{\ss} has formally defined KG as an acquisition and integration of information into ontology with a reasoner to derive new knowledge \cite{ehrlinger2016_KGDef}. 

In this study we consider WordNet as our use case KG. WordNet is an English words lexicon that organizes concepts into a conceptual hierarchy. It was designed to semantically model English words through the categorization of synonyms and existing taxonomic and non-taxonomic relations \cite{miller1995wordnet}. Since the creation of WordNet, it has become a valuable resource used in many domains, including Natural Language Processing (NLP), Information Retrieval (IR), and semantic-based recommender systems. Its semantic structure triggered the research community to examine tasks such as the one we investigate in this paper, namely semantic similarity and relatedness, with the objective of determining the level of likeness and connectedness between two entities based on their relations with other entities. While semantic similarity is a specific measure of likeliness, relatedness is a more general measure that also reflects connectedness. Semantic similarity and relatedness can be applied to solve challenging tasks, such as word sense disambiguation, text classification, information retrieval, machine translation, and document clustering.\cite{cai2018hybrid}

A wide range of semantic similarity measures have been proposed and applied in various applications and domains. These measures vary in performance based on their approaches and application domains. Detailed comparisons of these measures can be found in \cite{chandrasekaran2020evolution,cai2018measuring,cai2018hybrid,lastra2015,taieb2014,survey2014,lastra2019reproducible}. In summary, semantic similarity measures can be categorized into four main categories based on their approach: path, feature, Information Content (IC), and hybrid. 
Path-based measures count the number of edges in the shortest path between concepts. The longer the path between two concepts, the less similar they are, and vice versa\cite{zhu2016Computing,sebti2008,zhou2008SS}.  
Other Feature-based measures represent a concept as a vector of features constructed from its attributes \cite{zhu2017sematch,sanchez2012ontology,yang2019semantic}. Jiang et al. formally represented Wikipedia concepts as a structured knowledge base and proposed a multi-vector feature-based approach that includes features from concept's synonyms, glosses, Anchors, and Categories \cite{jiang2015feature}. Recent methods in this category incorporated Neural Network models to embed feature vectors that represent the entities, relations, and the entire Knowledge Graph (KG). These methods are being referred to as Knowledge Graph Embedding (KGE) \cite{camacho2016nasari,yu2019pykg2vec}. 
IC-based measures can be either extrinsic IC or intrinsic IC. Extrinsic IC measures are corpus-based, meaning an external corpus and a statistical model are used to compute the information of each concept
\cite{pedersen2004wordnet}. 
Intrinsic IC measures, on the other hand, are KG-structure-based, meaning the concept's information lies within the KG topological structure. Various structural attributes have been used as indicators of the information contained within each concept \cite{cai2018hybrid,zhang2018,meng2012,sanchez2011IC,zhou2008IC, sebti2008,seco2004}. Finally, hybrid similarity measures combine two or more of the above \cite{cai2018measuring,pirro2009,sebti2008,yang2019semantic}.

As mentioned above, an overwhelming number of semantic similarity measures have been proposed in the literature. Some researchers considered similarity to be a specific case of relatedness \cite{liu2012,cai2018hybrid}, while others did not distinguish between semantic similarity and relatedness \cite{seco2004, sebti2008}. Nonetheless, for the majority of these methods, similarity has been evaluated strictly based on hierarchical relations (i.e., hyponym/hypernym), with the exception of a few methods that have exploited a limited number of non-taxonomic relations, such as meronymy/holonymy and antonymy, to compute various relatedness measures \cite{sanchez2011IC, cai2018hybrid, zhou2008SS}.
However, WordNet includes many categories of non-taxonomic relations that have not yet been exploited by researchers to enhance semantic similarity and relatedness measures. For instance, those methods which used meronymy relation did not distinguish between the three types of meronymy relations: part of, substance of, and member of. This distinction is important because each meronym relation contributes different type of information to the semantic definition, and conveys specific information about the nature of the association between concepts. Based on the relation type (part of, substance of, and member of), the subject of an axiom conveys its inclusion in a larger entity, its physical components, or its participation in a group, respectively.
Furthermore, other non-taxonomic relations that have not yet been exploited so far, such as synonym, derivation, antonym, theme, cause, and action, convey an important informative component of a concept's semantic definition and information content. Therefore, a comprehensive semantic similarity measure should fairly incorporate information from all relations. Furthermore, relatedness in the literature was mostly evaluated based on the length of the path between two terms, without considering the importance of a particular relation within the modeled domain. We strongly believe that the most frequently used relations in a modeled domain, convey contextually related concepts in that domain. 

The main goal of this research is to demonstrate the importance of the information contained within non-taxonomic relations in enhancing the semantic similarity and relatedness between two terms. To achieve this goal, we propose a comprehensive poly-relational approach that exploits all non-taxonomic relations, their types, and their frequency, in evaluating semantic similarity and relatedness. In particular, the objectives of this research are:
\begin{itemize}
\item Address the limitations in existing IC-intrinsic and path-based similarity methods.
\item Demonstrate the importance of various non-taxonomic relations in enhancing semantic similarity and relatedness in the context of WordNet KG.
\item Combine semantic similarity and relatedness into a single comprehensive measure that complements existing taxonomic measures through the exploitation of all non-taxonomic relations embedded in WordNet.\item Evaluate our approach using gold standard semantic similarity benchmarks.
\item Compare our approach to the state of the art methods to demonstrate its robustness and scalability.
\end{itemize}

The rest of the paper is organized as follows: Section \ref{sec:related work} describes in details related work, motivations behind this study and contributions. Section \ref{sec:Proposed method} introduces the proposed method for developing a comprehensive semantic similarity and relatedness measure. In Section \ref{sec:Evaluation and Experimental Results}, we describe the experimental environment and discuss experimental results. Finally, conclusions are drawn and future research work is suggested in Section \ref{sec:Conclusion}.
\section{Related work}\label{sec:related work}
\begin{figure}[t]
	\centering
	\begin{subfigure}[b]{0.4\textwidth}
		\centering
		\includegraphics[height=3in,width=\textwidth]{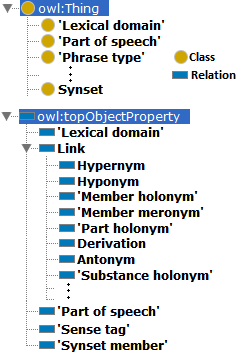}
		\caption{Sample ontology}
		\label{fig:wn_ontology}
	\end{subfigure}\qquad
	\begin{subfigure}[b]{0.4\textwidth}
		\centering
		\includegraphics[height=3in,width=\textwidth]{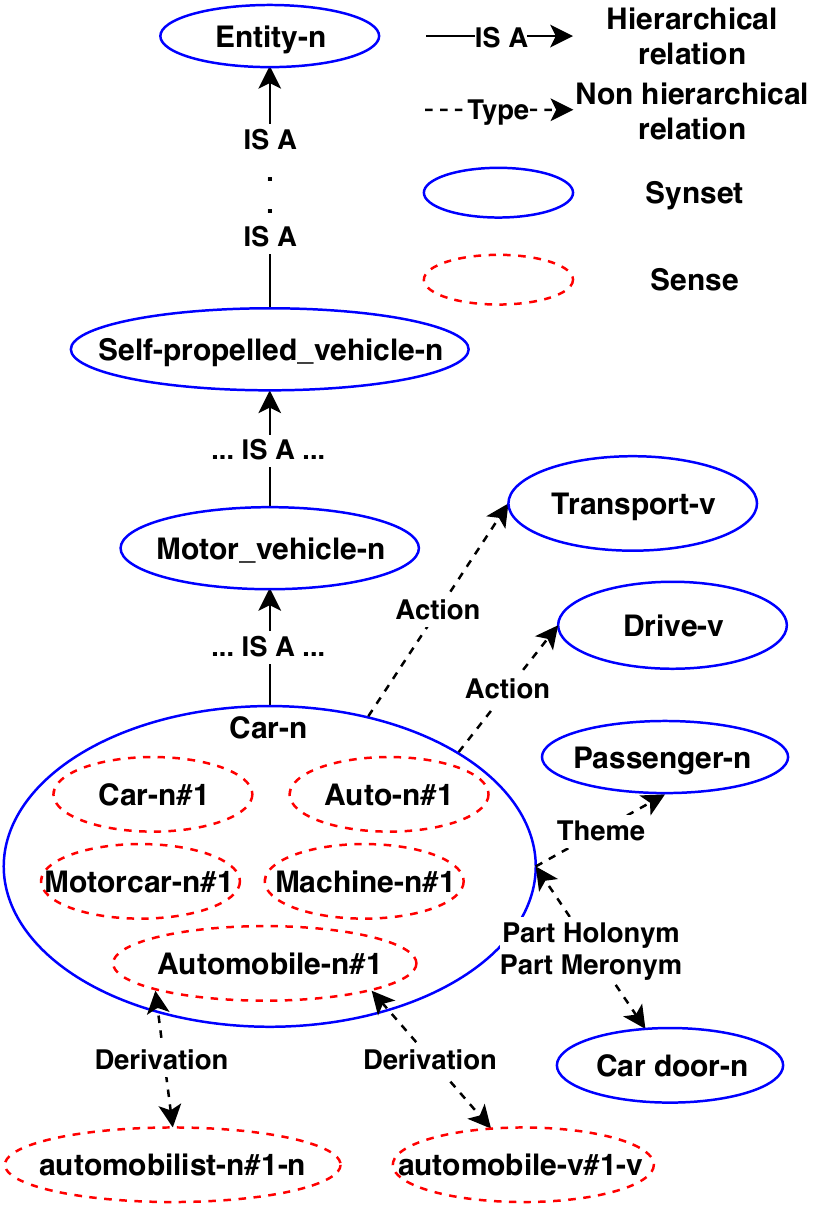}
		\caption{Sample graph}
		\label{fig:wn_graph}
	\end{subfigure}
	\caption{A fragment of WordNet ontology and graph}
	\label{fig:wn}
\end{figure}

In this section, we mainly focus on most prominent Intrinsic-IC based semantic similarity methods. Moreover, we present their main limitations with respect to the KG structure using the WordNet case study. 
Fig. \ref{fig:wn_ontology} describes the WordNet ontology that includes concepts and object properties, while Fig. \ref{fig:wn_graph} shows a KG representation of WordNet described in terms of instances of concepts and their relations based on the designed ontology and inferred knowledge.

\subsection{Intrinsic IC approaches}\label{subsec:Intrinsic IC Approaches}
This section presents the most commonly used IC-intrinsic methods, as well as their associated similarity measures. These are used as benchmarks to evaluate the method proposed in this paper. Table \ref{tb:icMitrics} lists the IC-intrinsic measures implemented through various topological features of WordNet.

Seco \cite{seco2004} was the first to introduce an intrinsic IC measure that is not dependent on an external corpus. His approach relies on the intrinsic features of the KG, specifically the number of hyponyms within the concept. He proposed an IC-intrinsic measure as a monotonically decreasing function with the number of hyponyms for a given concept. Seco's IC-intrinsic model proved that the number of hyponyms inversely conveys the concept's IC \cite{seco2004}. 

Using Seco's IC, Zhou \cite{zhou2008IC} and Cai \cite{cai2018hybrid} incorporated the concept's depth to emphasize the generalization/specialization effect on IC. Both approaches used depth to overcome Seco's method's limitation of attributing concepts with equal IC values regardless of their hierarchical level in the taxonomy. Zhou introduced a new IC measure as a function of normalized depth and hyponyms to compute the concept's IC \cite{zhou2008IC}, Cai proposed a new IC measure as a nonlinear transformation function to measure the contribution of depth to the concept's IC. Furthermore, Cai proposed a similarity measure to evaluate the IC measure \cite{cai2018hybrid}. 
  
Sebti proposed a new IC-intrinsic measure as a monotonically-increasing function of depth and number of siblings. He utilized the branching factor of all subsumers as an indicator of information gained through ancestor concepts. Hence, his new measure incorporated the number of subsumers with the probability of branching using direct hyponyms. Sebti's measure is not normalized, hence IC values could be $[0,\infty)$. Furthermore, he improved his IC metric with an edge counting-based tuning semantic similarity function \cite{sebti2008}. This approach clearly confirms the parent-child effect on IC, following the inheritance principle, while being a monotonically decreasing function moving from leaf to root.

Inspired by information theory, S\'anchez proposed a new leaf-based intrinsic IC metric. He argued that the concept's IC is directly proportional to its subsumers and inversely proportional to its leaves. Hence, his IC metric is described as a measure of concept's concreteness level to its abstraction level, specificity to generality. Unlike previous studies, S\'anchez incorporated multiple inheritance in the semantic similarity measure through the number of subsumers \cite{sanchez2011IC}.

In another study, Meng exploited the depth of a concept, as well as that of its hyponyms, in order to overcome Seco's approach of attributing the same IC value to all leaves. In other words, Meng's main argument was that leaves at a higher level of the taxonomy (i.e., smaller depth) convey less information than deeper leaves; hence, they have a smaller IC value \cite{meng2012}. 

Zhang introduced a new IC-intrinsic measure exploiting multiple inheritance. Zhang's improvement came from covering multiple inheritance concepts as well as incorporating the concept's siblings with depth, hyponyms, and hypernyms \cite{zhang2018}. 
Another Multiple inheritance approach was recently developed by Hussain \cite{hussain2020approach}. His approach utilizes a new neighbourhood ancestor semantic space to define concept's IC value. This technique is applied on a semi-structured taxonomy KG called Wikipedia Concept Graph (WCG) \cite{hussain2020approach}. 

Table \ref{tb:icMitrics} lists the IC-intrinsic measures described above, which are implemented through various topological features of WordNet. These IC measures were evaluated using either existing similarity measure such as Resnik
, Lin
, and JC 
\cite{pedersen2004wordnet}, or new similarity measure proposed by the authors. For example, Cai and Zhang \cite{cai2018hybrid,zhang2018} proposed new similarity measures exploiting the benefits of their new IC. Table \ref{tb:SimMetric} lists all pre-existing and new similarity measures based on IC-intrinsic measures. 

\begin{table}[!t]
\renewcommand{\tabcolsep}{0.02cm}
	\caption{ IC-intrinsic measures}\label{tb:icMitrics}
	\resizebox{\textwidth}{!}{
		\begin{tabular}[t]{p{.18\columnwidth}p{.75\columnwidth}p{.2\columnwidth}}
			\hline
			IC Measures & Formulae & Hierarchical features\\
			\hline
			Seco \cite{seco2004} & $ic_{seco}(c) = 1 - \frac{log{(hypo(c)+1)}}{log{(max_{wn})}} $ & hyponyms\newline \\ 
			Zhou \cite{zhou2008IC} & $ ic_{zhou}(c) = k\left(1 - \frac{log{(hypo(c)+1)}}{log(max_{wn})}\right) + (1-k)\left(\frac{log{(deep(c))}}{log(max_{deep})}\right) $ & hyponyms, depth\newline \\ 
			Sebti\protect\footnotemark \cite{sebti2008} & $ ic_{sebti}(c) = - log{\prod\limits_{c_i \in hyper(c)}^{} \frac{1}{DirHypo(c_i)}} $ & hypernyms, \newline direct hyponyms\\ 
			Meng \cite{meng2012} & $ ic_{meng}(c) = \frac{log{(deep(c))}}{log(max_{deep})} \times \Bigg (1-\frac{log\big (\sum\limits_{a\in hypo(c)}^{}\frac{1}{deep(a)}+1 \big )}{log(max_{wn})}\Bigg ) $ & depth, \newline hyponyms' depth\\ 
			S\'anchez \cite{sanchez2011IC} & $ ic_{s'anchez}(c) = - log{\bigg ( \frac {\frac{|leaves(c)|}{|subsumers(c)|}+1}{max\_leaves_{wn}+1} \bigg)} $ & leaves, \newline hypernyms\\ 
			Cai \cite{cai2018hybrid} & $ ic_{cai}(c)  = \left(1 - \frac{log{(hypo(c)+1)}}{log(max_{wn})}\right)\times tanh(deep(c)) $ & hyponyms,\newline  depth \\ %
			Zhang \cite{zhang2018} & $ ic_{zhang}(c) = K\left(1-\frac{log(hypo(c)+1)}{log(max_{wn})} \right)-(1-K) \frac{1}{n} \sum\limits_{i=1}^{n}log\left(\omega\right)$ & hyponyms, \newline hypernyms' siblings \\ 
			&\hspace{1cm} where \hspace{.5cm} $\omega = \prod\limits_{c_i\in hyper(c)}^{}\frac{1}{sibling(c_i)}+1$, \\
			&\hspace{2.4cm} $K = \frac{hypo(c)}{hyper(c) + hypo(c)}$, \\
			&\hspace{2.4cm} where $n $ is number of direct parents \\ 
			\hline
		\end{tabular}%
		}
\end{table}
\begin{table}[!h]
	\centering
	\caption{IC-based similarity measures}\label{tb:SimMetric}
		\begin{tabular}{ll}
			\hline
			Sim Measures & Formulae \\
			\hline
			Resnik \cite{pedersen2004wordnet}
			& $ sim_{res}(c_1,c_2) = \underset{c \in S(c_1,c_2)}{max} ic(c) $ \\
			Lin \cite{pedersen2004wordnet}
			& $ sim_{lin}(c_1,c_2) = \dfrac{2 \times sim_{res}(c_1,c_2)}{(ic(c_1) + ic(c_2))} $	\\
			JC \cite{pedersen2004wordnet}
			&	$ sim_{jc}(c_1,c_2) = 1 - \dfrac{ic(c_1) + ic(c_2) - 2 \times sim_{res}(c_1,c_2)}{2} $ \\
			Cai \cite{cai2018hybrid} & $ sim_{Cai_1}(c_1,c_2) = \exp^{-(\alpha\times spl_W(c_1,c_2)+\beta\times spl_N(c_1,c_2))} $ \\
			& $ sim_{Cai_2}(c_1,c_2) = \exp^{-(\alpha\times spl_W(c_1,c_2)+\beta\times spl_O(c_1,c_2))} $ \\
			& \hspace{10pt} $ spl_W(c_1,c_2) = ic(c_1) + ic(c_2) - 2 \times ic(lcs(c_1,c_2)) $ \\
			& \hspace{10pt} $ spl_N(c_1,c_2) = \frac{len(c_1,c_2)}{2\times Max_{deep}}$ \\
			& \hspace{10pt} $ spl_O(c_1,c_2) = log\Big(\frac{deep(c_1)+deep(c_2)+1}{2\times dep(LCS(c_1,c_2))+1}\Big)$ \\
			Zhang \cite{zhang2018} & $sim_{zhang}(c_1,c_2) = 1-log\Big(2-\frac{2\times ic(LCS(c_1,c_2))}{ic(c_1)+ic(c_2)} \Big)$ \\
			\hline
		\end{tabular} 	
\end{table}

\footnotetext{The authors did not explicitly state the final equation in their article. However, they demonstrated it through an example as follow: $ IC(Box)=-Log\left(\frac{1}{9} \times \frac{1}{10} \times \frac{1}{36} \times \frac{1}{42} \times \frac{1}{13} \times \frac{1}{49}\right)=18.2778  $, where the denominator represents the number of siblings from the highest subsumer to the concept}
\subsection{Non-taxonomic approaches}\label{subsubsec:Non-Taxonomic Approaches}
Few studies used non-hierarchical relations to solve challenges such as spelling errors correction and Word Sense Disambiguation (WSD) based on the relatedness between concepts. In \cite{pesaranghader2013}, the authors attempted to solve WSD by using a gloss vector. They evaluated a relatedness measure between concepts using standard vector-based similarity measures (i.e., overlap, cosine similarity). The vector's dimensions are words extracted from a concept's glossary in WordNet. However, the employed method was more of a linguistic/NLP approach rather than a semantic one, as they evaluated the English definition from a glossary rather than semantic relations.

In Liu \cite{liu2012}, concepts are expressed by their relevant concepts as a vector. Relevancy is defined by the set of hypernyms and hyponyms. Dimensions are represented as their local densities (i.e., number of siblings in this case). To improve similarity and relatedness, the authors computed the relatedness strength between two concepts based on the number of paths between them. A path could be direct PartOf path (i.e., one concept is part of the other), or indirect (i.e., one concept is part of an element from the relevant set of the other). The relatedness strength is then added to the LCS as a new sibling, and a taxonomic-based approach is applied to compute similarity \cite{liu2012}. A major limitation in this approach is that the paths are not pure relational, but mostly hierarchical. It nonetheless demonstrated that multiple paths are directly proportional to the relatedness strength, and can be employed to improve relatedness.

In \cite{cai2018measuring}, the authors explored a new path-based approach. $Path_{ISA}$ and $path_{PartOf}$ are computed based on ISA and PartOf relations, respectively, and the shortest of the two was selected to compute the similarity level between the two concepts. The main innovation of this method lies within the $path_{ISA}$ taxonomic approach. This is because the $path_{PartOf}$ is limited to a direct PartOf relation, or at most two such relations that connect two concepts through a common meronym \cite{cai2018measuring}. This is a major limitation of this approach, especially given that no (or few) pairs with such paths exist in the used datasets. 

\begin{table}
    \centering
	\caption{Non-taxonomic semantic relations in WordNet}
	\label{tb:wn_RelationLinksSyn}
	\resizebox{.75\textwidth}{!}{
		\begin{tabular}{p{.4\columnwidth}>{\raggedleft}m{.25\columnwidth}r}
			\hline
			\textbf{Relation Name} & \textbf{Frequency} & \textbf{Prevalence} \\
			\hline 
			\textbf{synset\_member (synonym) } 	&	145076	&	74.61\% \\
			\textbf{member\_meronym	} 	&	12252	&	6.30\% \\
			\textbf{member\_holonym	} 	&	12242	&	6.30\% \\
			\textbf{part\_meronym	} 	&	9082	&	4.67\% \\
			\textbf{part\_holonym	} 	&	9071	&	4.67\% \\
			\textbf{derivation	} 		&	2957	&	1.52\% \\
			\textbf{antonym	} 			&	2154	&	1.11\% \\
			\textbf{substance\_holonym} &	746		&	0.38\% \\
			\textbf{substance\_meronym} &	744		&	0.38\% \\
			\textbf{theme	} 			&	103		&	0.05\% \\
			\textbf{cause	} 			&	15		&	0.01\% \\
			\textbf{action	} 			&	3		&	0.00\% \\
			\hline
		\end{tabular}
	}
\end{table}

\subsection{Motivations and contributions}\label{subsec:Motivations and Contribution}

\begin{figure}[!t]
    \centering
    \begin{subfigure}[b]{.35\textwidth}
        \includegraphics[width=.5\textwidth,height=.5in]{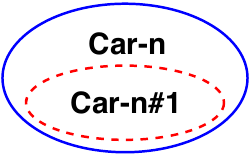}
        \caption{Car keyword}
    	\label{fig:RE_car}
        \vspace{3em}
        \includegraphics[width=\textwidth,height=1.25in]{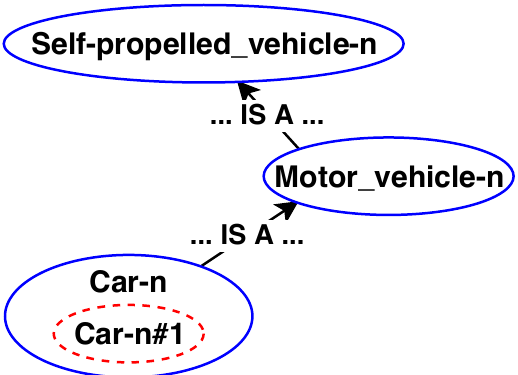}
        \caption{Car with taxonomic relations}
    	\label{fig:RE_ISA}
    \end{subfigure}\qquad
    \begin{subfigure}[b]{.5\textwidth}
        \includegraphics[width=\textwidth,height=2.5in]{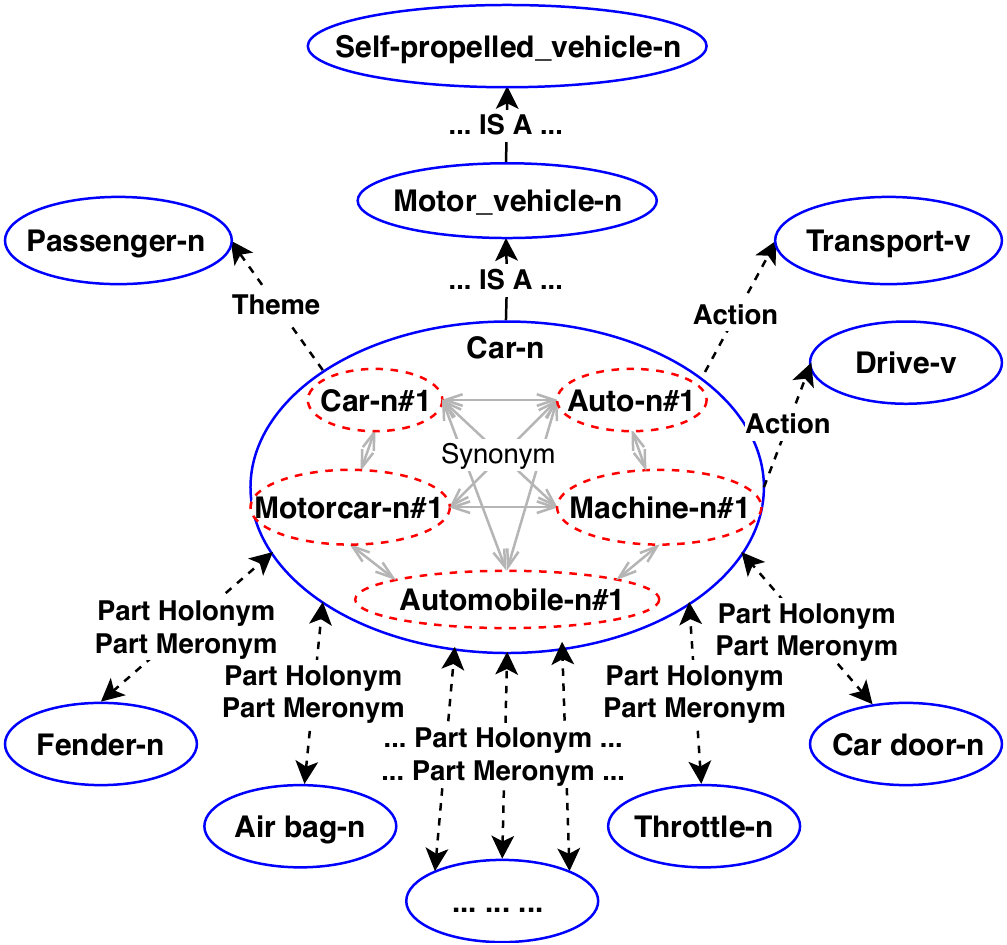}
        \caption{Car with all semantic relations}
		\label{fig:RE_relations}
    \end{subfigure}
    \caption{The car concept in WordNet graph}
	\label{fig:RelEffect}
	\vspace{-.5cm}
\end{figure}

Various intrinsic IC measures were proposed and used to determine the semantic similarity between concepts. As described in Section \ref{subsec:Intrinsic IC Approaches}, these IC measures exploited different taxonomic features of KG. A common limitation of these measures is that they rely solely on a single semantic dimension --- the taxonomic "is a" (ISA) relation --- and that they ignore all other semantic dimensions, hence, limiting semantic similarity strictly to the generalization/specialization relation. However, by definition, \textit{"Semantics give a keyword symbol useful meaning through the establishment of relationships"} \cite{hebeler2011SemanticWeb}. 
This is clearly illustrated in the example shown in Fig. \ref{fig:RelEffect}. 
The isolated first sense of the noun \textit{Car}, denoted as \textit{Car-n\#1}, has no semantic meaning except that it is a member of synset \textit{Car-n} (see Fig. \ref{fig:RE_car}). When adding the taxonomic semantic relation ISA, as shown in Fig. \ref{fig:RE_ISA}, one can then elaborate on the definition as follows: \textit{"Car is a motor vehicle, which is a self-propelled vehicle, etc."}. However, when we further include explicit (e.g. \textit{part holonym /meronym, theme, action, etc.}) and implicit (i.e. synonym) non-taxonomic relations, the semantic definition of \textit{Car-n\#1} is significantly enriched, and hence its IC. The new definition of \textit{Car-n\#1} would be: \textit{"Car is a motor vehicle, which is a self-propelled vehicle. Car has a theme as passenger. Car leads to actions such as transport and derive. Car has some parts such as car door, throttle, air bag, fender, etc. Car has synonyms (auto, machine, motorcar, and automobile)"}. To overcome this limitation, we argue that semantic relations other than ISA enhance the concept's semantic definition and increase its IC value; an increase that is proportional to the type and strength of that relation. Hence, this work extends the taxonomic-based IC principle applied in previous literature by exploiting non-taxonomic semantic relations.

\begin{figure}[t]
	\centering
    \begin{subfigure}[b]{0.45\textwidth}
    	\includegraphics[width=\textwidth,height=2.5in]{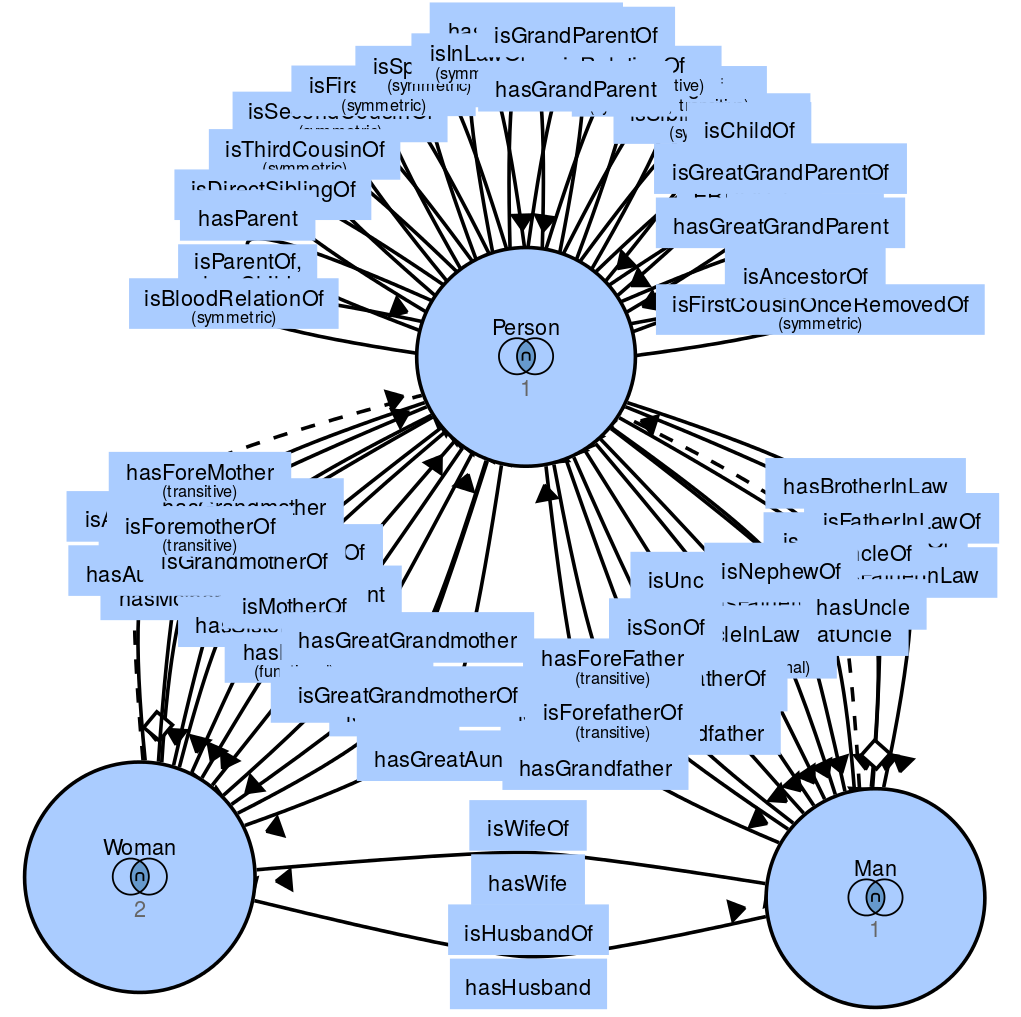}
    	\caption{Segment of ontology\protect \footnotemark}
    	\label{fig:familyOntology}
	\end{subfigure}
    \begin{subfigure}[b]{0.45\textwidth}
        \includegraphics[height=2.5in]{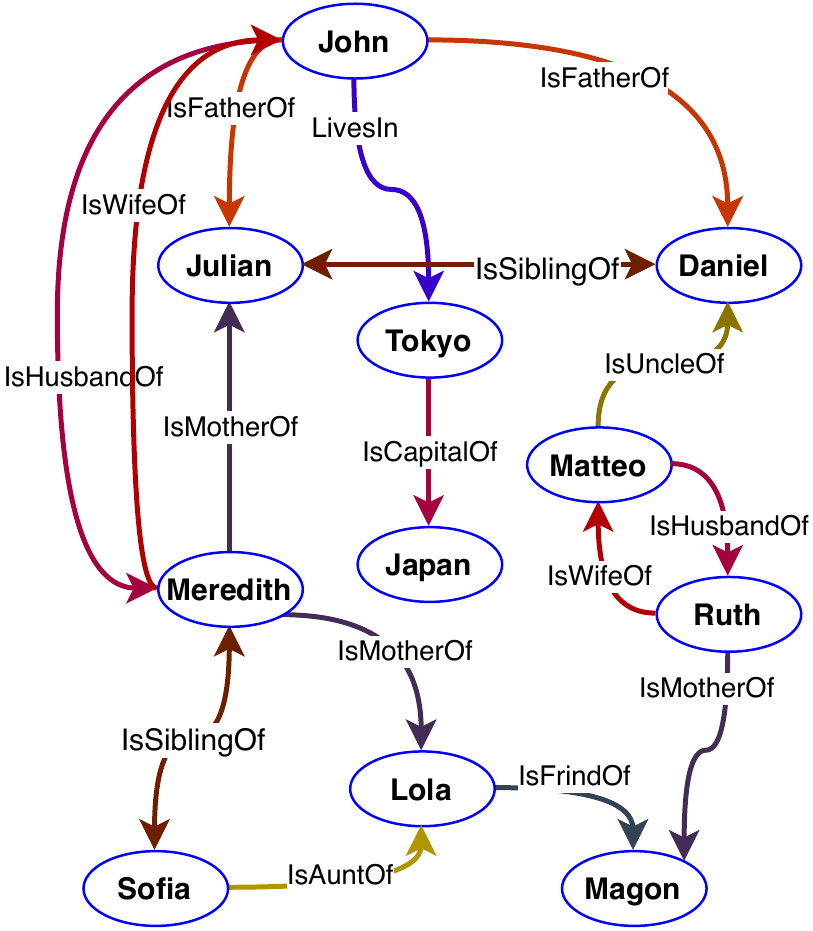}
        \caption{Segment of KG}
		\label{fig:family_KG}
    \end{subfigure}
    \caption{Segments of the family ontology and KG}
	\label{fig:FamilyOnt_KG}
\end{figure}

As described in Section \ref{subsubsec:Non-Taxonomic Approaches}, it can also be observed that the few non-taxonomic approaches available in the literature do not exploit non-taxonomic relations to their full potentials. They either limit the non-taxonomic relational path to one or two links, or rely only on a single relation (i.e. PartOf). Domain ontologies and KGs are contextually designed for specific purposes, therefore they include a rich set of non-taxonomic relations to contextually describe the relationships between concepts. This is clearly illustrated in Fig. \ref{fig:familyOntology}, which describes the family ontology in terms of  relations that model real world family relationships such as husband, child, spouse, and sibling (see Fig. \ref{fig:familyOntology}). Fig. \ref{fig:family_KG} on the other hand describes an example KG of family domain with many non-taxonomic relations describing various relationships between persons of the modeled family. Relatedness between persons of a family can be better evaluated using these relations, focusing on their types and occurrences in a path between two persons. Therefore, to overcome the limitations of existing non-taxonomic approaches, in this research, we take into consideration the type and frequency of all non-taxonomic relations to measure relatedness between concepts. Thus, incorporating all contextually related concepts and privileging the dominant relations in the described domain. Consequently, in this paper we propose a weighted relational-based path approach, exploiting all non-taxonomic relations between two concepts to measure their relatedness and enhance their semantic similarity.

To summarize, various principles and findings from the above-mentioned analysis of the literature motivated this research. Firstly, the role of relationships in ontology and KG design, and their importance to serve a contextual purpose within the modeled domain. Ontologies are domain-specific with a precise set of relations that enhances the semantic representation of data in the KGs \cite{hebeler2011SemanticWeb}. Therefore, ignoring these relations leads to an incomplete semantic evaluation. Secondly, our approach takes into consideration the information inheritance principle, such as the one used in the taxonomic approaches based on ISA relation. A child concept accumulates information from its parent concept(s), and adds its own new information. However, we strongly believe that other relations also contain and convey important information between concepts. Therefore, the effect of non-taxonomic relations on similarity is unavoidable. Finally, it should be noted that the benchmark datasets are measured by humans based on the combination of similarity and relatedness. Therefore, to provide fair evaluation and comparison with the gold standard, our approach makes use of all types of relations in a KG to devise a single measure that incorporates both the semantic similarity and relatedness between concepts.
\footnotetext{Visualization is done using \url{"http://vowl.visualdataweb.org/webvowl.html"}}

Inspired by the related work, and motivated by the drivers put forward to this research, our research is based on the following observations:

\begin{obs} \label{obs:first}
Considering non-taxonomic relations in a KG enhances information content, thus yielding more semantic similarity between concepts than just relying on taxonomic relations.
\end{obs}
\begin{obs} \label{obs:second}
The prevalence of a non-taxonomic semantic relation within a KG is an indicator of its importance and relevance to the modelled domain. Thus, it has an impact on the relatedness between concepts.
\end{obs}
The contributions of this research are summarized in the following points:
\begin{itemize}
\item A new relation-weighting schema based on the IC difference between linked concepts for measuring non-taxonomic relation-based semantic similarity.
\item A new relatedness measure based on IC-weighted path(s) of non-taxonomic relations between terms.
\item A holistic poly-relational approach that exploits all non-taxonomic relations, their types, and their frequency, for enhancing semantic similarity and relatedness in the context of WordNet.
\end{itemize}

\section{Proposed method}\label{sec:Proposed method}
Motivated by the previous work and aforementioned observations, we propose a holistic poly-relational semantic similarity and relatedness measure. The proposed approach complements existing taxonomic measures through the exploitation of all non-taxonomic relations embedded in WordNet. In Section \ref{subsec:Relations IC}, we introduce the concept of relation IC, that is the information contained within a non-taxonomic relation. In Section \ref{subsec:Relations prevalence}, we investigate the contextual importance of relations by introducing the relation prevalence concept. Then, we present our poly-relational semantic similarity and relatedness method as a function of IC-based taxonomic similarity, non-taxonomic relation-based similarity, and weighted IC-based relatedness, exploiting the two main building blocks from Sections \ref{subsec:Relations IC} and \ref{subsec:Relations prevalence} to compute these measures in four different strategies. 

The first three strategies demonstrate the way non-taxonomic relations enhance semantic similarity at different granularity levels. In the first strategy we demonstrate similarity based on the term's semantic information content obtained from non-taxonomic relations. The second and third strategies are similarity based on the commonality of the relations’ types and their instances, respectively. In the fourth strategy, we show the usefulness of relatedness by further improving semantic similarity. We achieve that by introducing a new contextualized relatedness measure based on weighted IC values of all non-taxonomic paths. A detailed description of the proposed method and strategies is given below.

\subsection{Relation IC}\label{subsec:Relations IC}
Similar to concepts, relations are organized into hierarchical structure within their ontology. For instance, the ontology in Fig. \ref{fig:wn_ontology} shows a sub-set of the relations' hierarchy used in Wordnet, where \textit{topObjectProperty} is the most general relation, and \textit{Hypernym, Hyponym, and others} are some of the most specific relations in the hierarchy. Based on such hierarchy, concepts gained an intrinsic IC attribute, and likewise, do relations. Having this intuition, we utilize existing IC-intrinsic metrics with relations to compute the $ IC_{Tax}(r_t) $ as the taxonomic-based IC for relation $r$ of type $t$. Eq. \eqref{eq:Rel_IC_Tax} simply reflects the taxonomic IC for each relation type.
\begin{equation}\label{eq:Rel_IC_Tax}
    IC_{Tax}(r_t) = ic_{base}(r_t),
\end{equation}
where $ic_{base}$ denotes the baseline taxonomic IC measure from Table \ref{tb:icMitrics}.

Furthermore, domain and/or range concepts in an ontology may contextualize a relation. The IC of domain and range concepts enrich the associated relation as global contextual information. Intuitively, we argue that relations assimilate global contextual information from their domain and range concepts, which is proportional to their depths. Therefore, we define the global IC for a relation as a monotonically increasing function with respect to both the IC and depth of the domain and range concepts, as shown in Eq. \eqref{eq:Rel_IC_GC}. Consequently, for two pairs of concepts with the same absolute IC difference, the deeper the pair, the greater the global IC. Also, for two pairs of concepts at the same depth, the greater absolute IC difference, the greater the global IC. It should be noted here that Eq. \eqref{eq:Rel_IC_GC} is a monotonically increasing function with respect to both the IC and depth of the domain and range concepts, as shown below: 
\begin{equation}\label{eq:Rel_IC_GC}
    IC_{GC}(r_t)=\Big|ic_{base}\Big(Dom(r_t)\Big)-ic_{base}\Big(Ran(r_t)\Big)\Big|^{\varPsi},		
\end{equation}
where $ic_{base}(Dom(r_t))$, and $ic_{base}(Ran(r_t))$ are the ICs of the domain and range concepts respectively, and $\varPsi=[1/(deep(Dom(r_t))+deep(Ran(r_t)))]$, where $deep$ is the depth function. 

\begin{figure}[t]
	\begin{subfigure}[b]{.55\textwidth}
        \includegraphics[width=.95\textwidth]{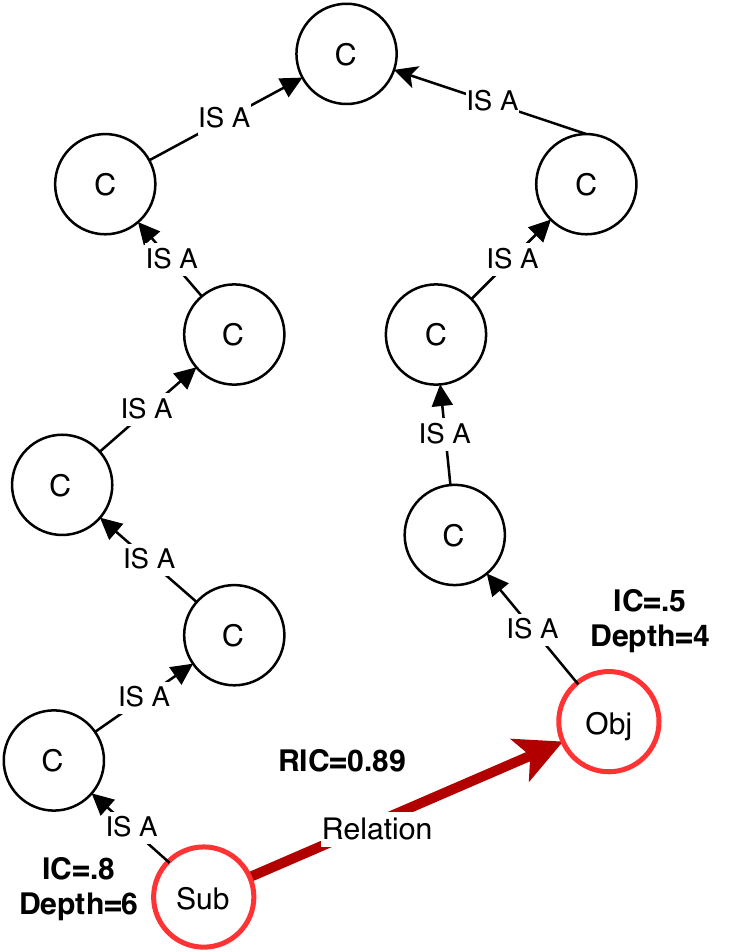}  
        \caption{Deep concepts}
        \label{fig.sub:RIC_depth_Effect:a}
	\end{subfigure}
	\begin{subfigure}[b]{.45\textwidth}
        \includegraphics[width=.95\textwidth]{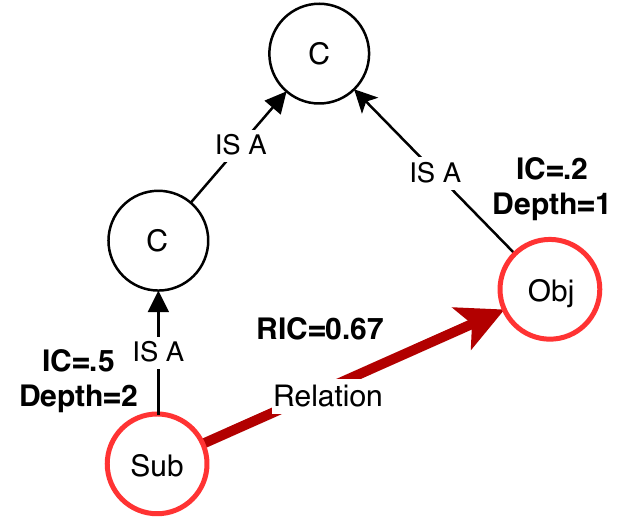} 
        \caption{Shallow concepts}
        \label{fig.sub:RIC_depth_Effect:b}
	\end{subfigure} 
	\caption{Concept's depth effect on RIC}
	\label{fig:RIC_depth_Effect}
\end{figure}

Similar to the global context, an instance relation attains information from its subject and object within the KG. Hence, local contextual information is assigned to each instance relation based on its subject and object. The local IC for a relation instance of type $t$ is defined in the equation below:
\begin{equation}\label{eq:Rel_IC_LC}
    IC_{LC}(r_t)= \Big|ic_{base}\big(Sub(r_t)\big)-ic_{base}\big(Obj(r_t)\big)\Big|^{\varUpsilon},
\end{equation}
where $Sub(r_t)$, and $Obj(r_t)$ are the subject and object concepts, respectively, and $\varUpsilon=[1/(deep(Sub(r_t))+deep(Obj(r_t)))]$. Fig. \ref{fig:RIC_depth_Effect} illustrates the effect of the depth of the concepts on the relations' IC values. This effect applies to both the global and the local ICs within the ontological schema and knowledge graph instances respectively. Fig. \ref{fig.sub:RIC_depth_Effect:a} shows a Relation Information Content (RIC) between deeper concepts while Fig. \ref{fig.sub:RIC_depth_Effect:b} illustrates an RIC between shallow concepts.

Finally, the taxonomic, global context, and local context IC values are combined to form the Relation Information Content ($RIC$), as shown in the following equation:
\begin{equation}\label{eq:Rel_RIC}
    RIC(r_t)= \alpha \times IC_{Tax}(r_t) +\beta \times IC_{GC}(r_t) +\gamma \times IC_{LC}(r_t),
\end{equation}
where $ \alpha $, $ \beta $, and $ \gamma $ are constants to measure the contribution of each IC function. These constants are contextually selected based on the actual conceptual schema and hierarchical structure of the ontology and KG describing the modeled knowledge.

\subsection{Relations prevalence}\label{subsec:Relations prevalence}
Based on the second observation described in Section \ref{subsec:Motivations and Contribution}, relations exist in a KG in accordance to their relevancy to the modelled domain. Hence, the probability of a relation in a KG, Eq. \eqref{eq:Rel_prob}, measures the contribution of that relation to the semantic information added to each associated concept. 
\begin{equation}\label{eq:Rel_prob}
    P(r_t)=\frac{freq(r_t)}{Total\, number\, of\, relations},
\end{equation}
where $freq(r_t)$ is the total number of relations of type \textit{t}. 
\subsection{Poly-Relational similarity and relatedness measure}\label{subsec:Poly-Relational similarity and relatedness measure}
\begin{figure}
    \centering
    \includegraphics[width=\textwidth]{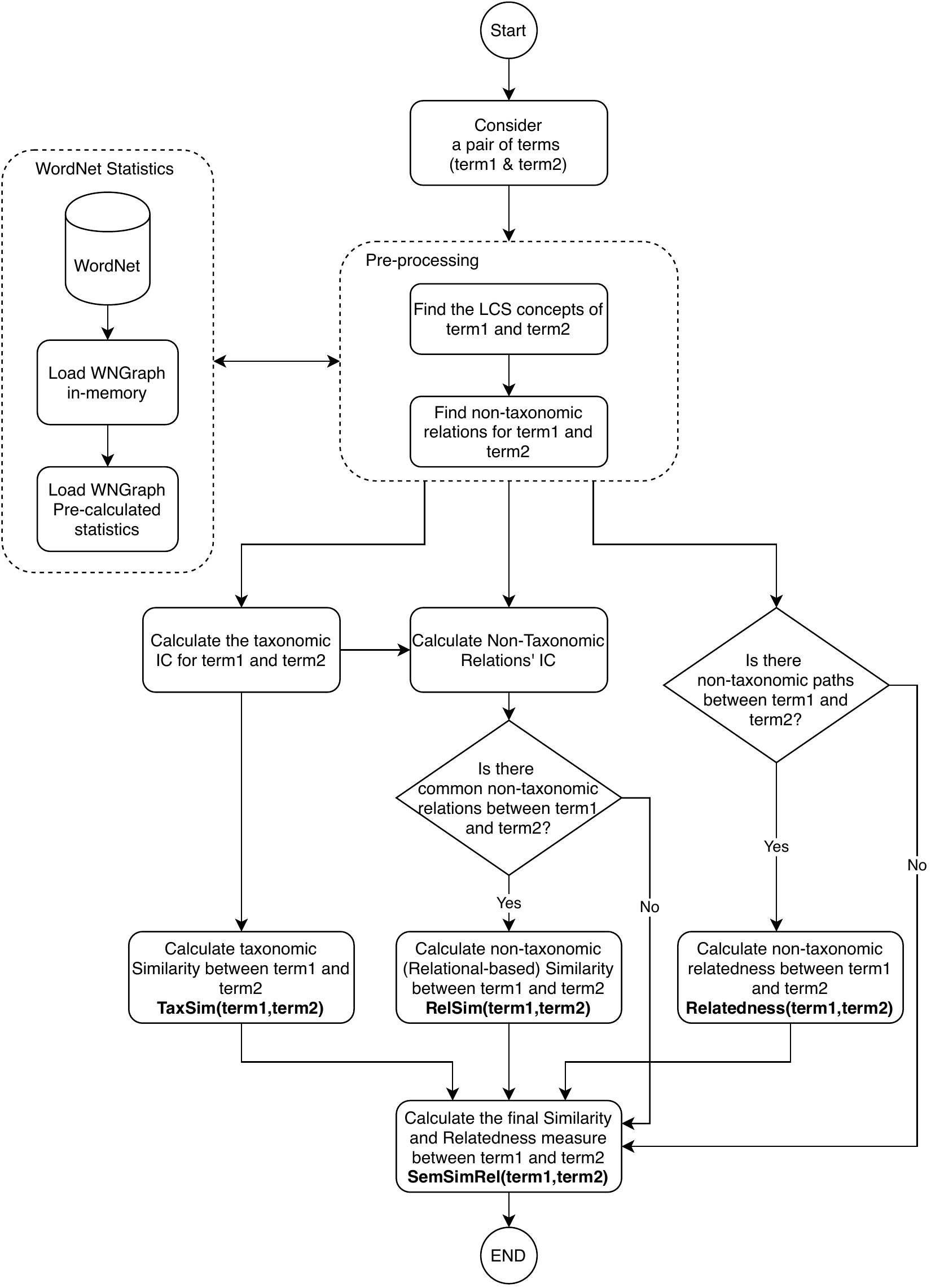}
    \caption{Flow chart for the semantic similarity and relatedness algorithm}
    \label{fig:SSR_Flowchart}
\end{figure}

We propose a semantic similarity and relatedness measure between two terms as a function of their taxonomic similarity, relational similarity, and relatedness.
\begin{equation}\label{eq:poly-rel}
\resizebox{.92\textwidth}{!}{$SemSimRel(w_1,w_2)= f\Big(TaxSim(w_1,w_2), RelSim(w_1,w_2), Relatedness(w_1,w_2)\Big),$}
\end{equation}
where $TaxSim(w_1,w_2)$ represents a general taxonomic IC-based similarity measure between $w_1$ and $w_2$ as described in Table \ref{tb:SimMetric}, $RelSim(w_1,w_2)$ is the proposed relational-based non-taxonomic similarity presented in Section \ref{subsec:Relational-base similarity}, and $ Relatedness(w_1,w_2) $ is the proposed relatedness based on weighted non-taxonomic relational paths presented in Section \ref{subsec:Relatedness}. Fig. \ref{fig:SSR_Flowchart} depicts the procedure to of calculating the semantic similarity and relatedness using the above equation. Unlike other methods, we combine existing taxonomic measures with the proposed non-taxonomic relational-based similarity and non-taxonomic relatedness measures between terms.

Below we show how relation IC and relation prevalence are employed to devise a novel comprehensive semantic similarity and relatedness measure. 

\subsection{Relational-based similarity}\label{subsec:Relational-base similarity}
In relational-based similarity, terms are considered similar based on the similarity of their semantic non-taxonomic relations. To demonstrate the benefit of employing relations to enhance semantic similarity, we propose three strategies showing their impact at different granularity levels. The first strategy makes use of all relations, regardless of their type, to compute a single semantic attribute. However, the second strategy exploits relations' type in computing similarity. Finally, the third strategy computes similarity based on instances of each relation type. This coarse-to-fine grain investigation provides new insights about the role of non-taxonomic relations in measuring semantic similarity. Furthermore, a fourth strategy based on non-taxonomic relational paths is proposed to bring into perspective the role of relatedness in further enhancing semantic similarity.

\subsubsection*{Strategy 1}\label{subsubsec:Strategy 1}
\begin{figure}[t]
    \centering
    \includegraphics[width=.9\textwidth]{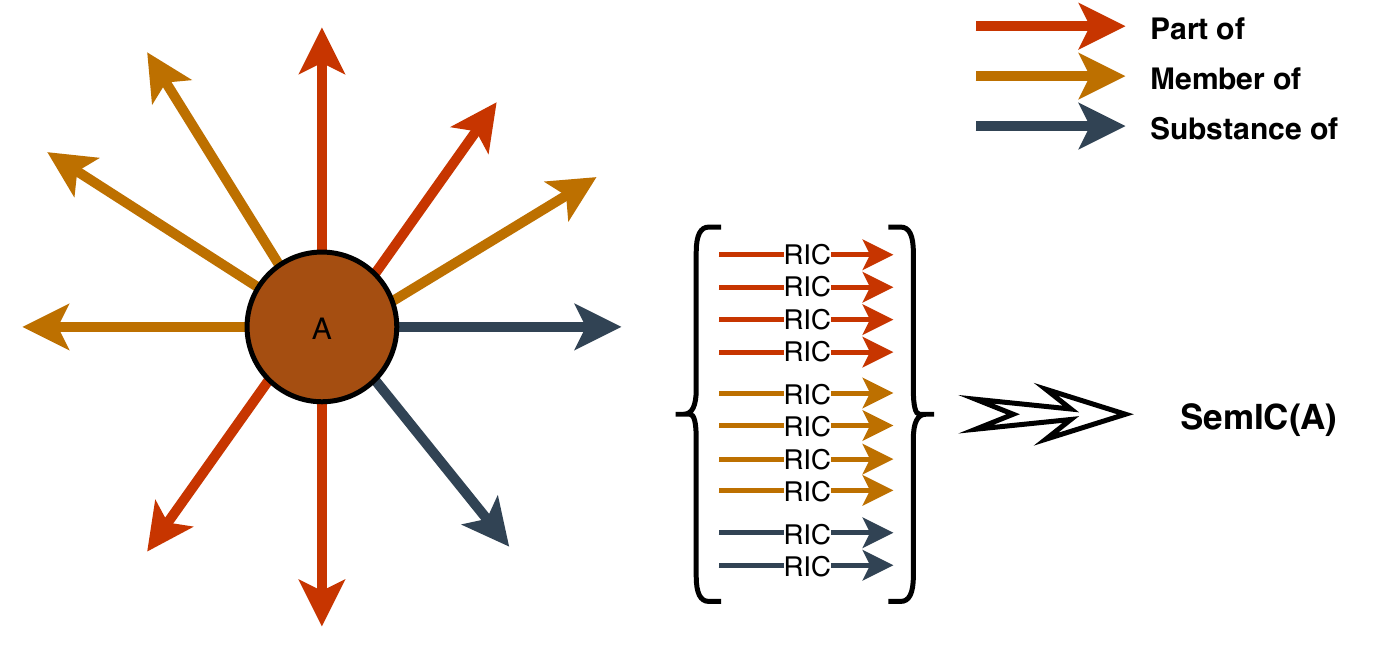}
    \caption{Semantic Information Content for concept based on Strategy 1}
    \label{fig:S1_SemIC}
\end{figure}

This strategy is based on the intuition of the original taxonomic IC, where each concept is attributed an IC value as a measure of various hierarchical features (i.e., hyponyms, depth, hypernyms, leaves, etc.) as shown in Table \ref{tb:icMitrics}. Similarly, we propose an additional semantic IC (\textit{SemIC}) attribute for each term, based on all associated non-taxonomic relations. This semantic IC metric, defined in Eq. \eqref{eq:S1_SemIC}, is an aggregation of the RIC contribution of all associated relations. 
\begin{equation}\label{eq:S1_SemIC}
    SemIC(w)= log\left(\sum\limits_{r_t\in RelSet:\left(w\overset{r_t}{\rightarrow}o\right)}\hspace{-.75cm}\left(P(r_t)\times RIC(r_t)\right)+1\right),
\end{equation}
where $RelSet(w\overset{r_t}{\rightarrow}o)$ is the set of all relation instances that link term $w$ with all other object terms. Fig. \ref{fig:S1_SemIC} illustrates a concept semantic IC (SemIC) as a single value computed using Eq. \ref{eq:S1_SemIC}.

The semantic IC then used to evaluate the semantic similarity between terms, using existing baseline similarity measures from Table \ref{tb:SimMetric}. The relational similarity $RelSim(w_1,w_2)$ between $w_1$ and $w_2$ is computed by replacing the $ic(w)$ by $SemIC(w)$, as denoted below:
\begin{equation}\label{eq:S1RelSim}
    RelSim(w_1,w_2)=Sim(w_{1_{SemIC}},w_{2_{SemIC}}),
\end{equation}
where $w_{i_{SemIC}}$ refers to the semantic IC of term $i$ instead of its taxonomic IC. 

\subsubsection*{Strategy 2}\label{subsubsec:Strategy 2}
\begin{figure}[t]
    \centering
    \includegraphics[width=.9\textwidth]{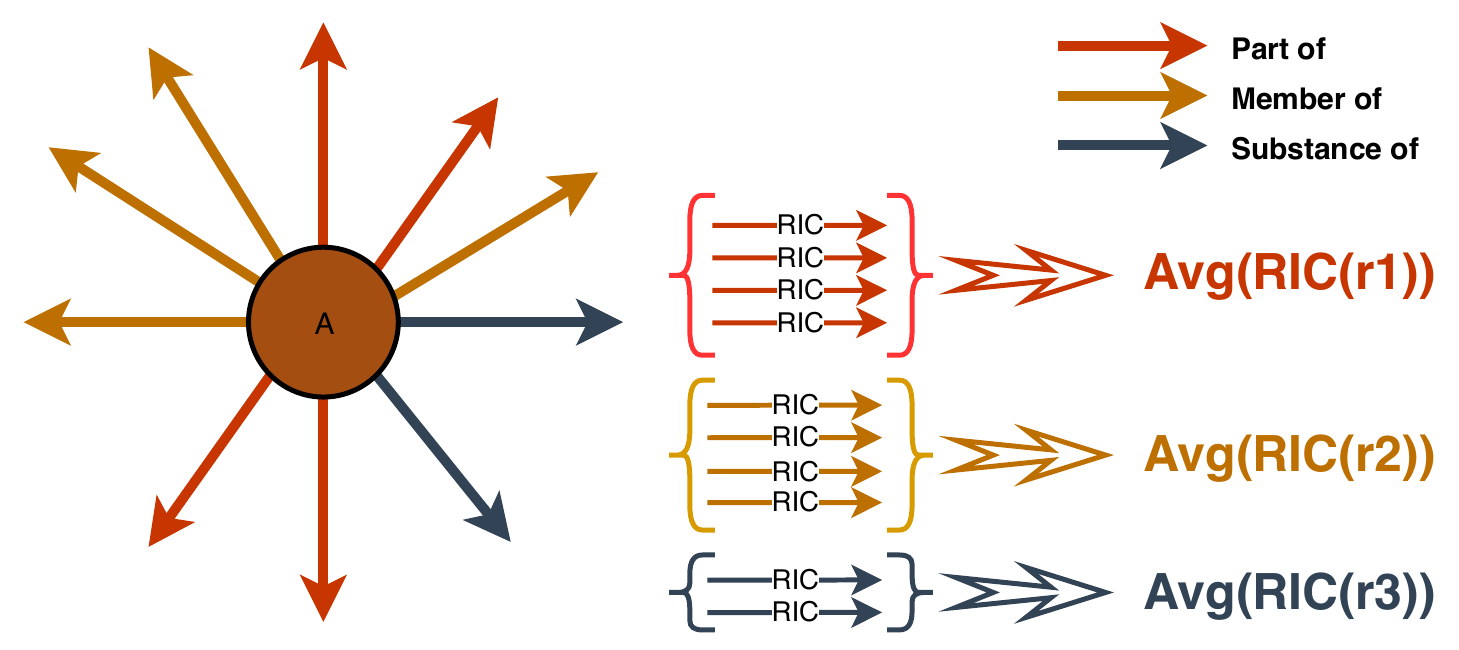}
    \caption{Semantic Information Content for concept based on Strategy 2}
    \label{fig:S2_SemIC}
\end{figure}

The non-taxonomic semantic IC, \textit{SemIC}, proposed in the first strategy assigns a non-discriminating semantic attribute to each concept. Although this attribute describes the semantic information contained within the concepts, yet, relations' types are being ignored in the similarity evaluation between two concepts.
Furthermore, in Section \ref{subsubsec:Non-Taxonomic Approaches}, we demonstrated the importance of each relation type and its own implication on the semantic definition and IC of a concept. Therefore, the second strategy refines the first one by attributing a vector of RICs for each term. Each element of the vector represents the average $RIC$ of a specific relation type. Fig.\ref{fig:S2_SemIC} illustrates three non-taxonomic relation types (part of, member of, and substance of). For each relation type, the average RIC is computed and used to build a final SemIC vector as shown in Eq. \ref{eq:S2_SemIC} below:
\begin{equation}\label{eq:S2_SemIC}
    SemIC(w)=\vec{r}_{w}=\begin{bmatrix}
                        Avg(RIC(r_{t1})) \\ 
                        ... \\ 
                        Avg(RIC(r_{tn}))
                        \end{bmatrix},
\end{equation}
where $r_{ti}\in RTSet(w)$ denoting the set of relation types linking the term $w$ to all other objects.

We then employ one of the existing vector-based similarity/distance measures, such as Euclidean, Hamming, Cosine, Mean-Squared Error (MSE), and Summation of Squared Difference (SSD), to measure relational similarity. We compared eleven distance measures from Math.NET Numerics library\footnote{\url{https://numerics.mathdotnet.com/Distance.html}}. 
\begin{equation}\label{eq:MSE}
    MSE(\vec{x},\vec{y})=\frac{1}{n}\sum_{i=1}^{n}{(x_i-y_i)^2},
\end{equation}
The MSE distance, described in Eq. \eqref{eq:MSE}, provided the highest correlation results with the gold standard, and consequently, is used to compute the relational similarity as shown in Eq. \eqref{eq:S2RelSim}
\begin{equation}\label{eq:S2RelSim}
    RelSim(w_1,w_2)=1-MSE(\vec{r}_{w_1},\vec{r}_{w_2}),
\end{equation}
where $ \vec{r}_{w_1} $ and $ \vec{r}_{w_2} $ are the average relational type vectors of $ w_1 $ and $ w_2 $, respectively.

\subsubsection*{Strategy 3}\label{subsubsec:Strategy 3}

\begin{figure}[t]
    \centering
    \includegraphics[width=.9\textwidth]{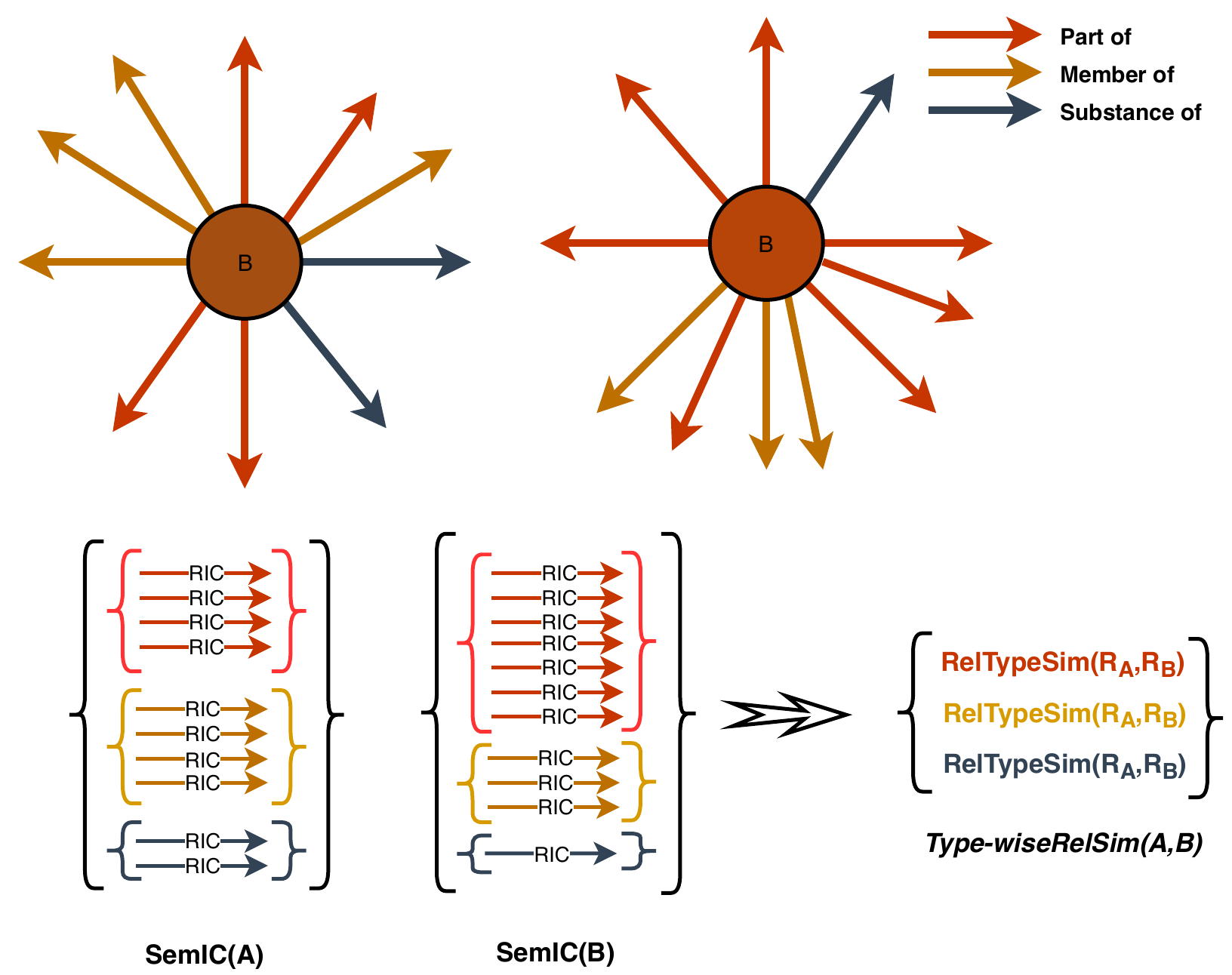}
    \caption{Semantic Information Content for concept based on Strategy 3}
    \label{fig:S3_SemIC}
\end{figure}
This strategy is more granular than the previous ones, where we focus not only on the types of relations as we did in strategy 2, but also on each instance within each relation type. More specifically, this approach exploits the benefits of measuring similarity between instances of the common relation types of both terms. Then, aggregating the resulting type-wise relations' similarities to compute the overall relational similarity.

As illustrated in Fig. \ref{fig:S3_SemIC}, first, for each relation type in the $RTSet(w)$, we generate a vector of RICs associated to the relation instances of that type. Then, these vectors are used to build a semantic IC of a term as a set of type-wise \textit{RIC} vectors denoted by \{$\vec{r}_{ti}, i=1..n$\}, where $n$ is the number of relation types associated to that term, as shown in Eq. \eqref{eq:S3_SemIC}.
\begin{equation}\label{eq:S3_SemIC}
    SemIC(w)=\left\{\begin{matrix}
                \vec{r}_{t1}=\begin{bmatrix}
                            RIC(r_{t11}) \\
                            ... \\ 
                            RIC(r_{t1m_1})
                \end{bmatrix}\\ 
                ...\\ 
                \vec{r}_{tn}=\begin{bmatrix}
                            RIC(r_{tn1}) \\ 
                            ... \\ 
                            RIC(r_{tnm_n})
                \end{bmatrix}
    \end{matrix}\right\},
\end{equation}
where $m_i$ is the cardinality of instances of the $i^{th}$ relation type. To compute the relational type-wise similarity between two terms, we use the $MSE$ distance method as shown below: 
\begin{equation}\label{eq:RelTypeSim}
    RelTypeSim(r_{t_{w_1}},r_{t_{w_2}})=1-MSE(\vec{r}_{t_{w_1}},\vec{r}_{t_{w_2}})
\end{equation}
Consequently, a type-wise relational similarity set for the common relation types between two terms can be expressed as follows:
\begin{equation}\label{eq:Type-wiseRelSim}
    Type\mbox{-}wiseRelSim(w_1,w_2)=\left\{\begin{matrix}
        RelTypeSim(\vec{r}_{t1_{w_1}},\vec{r}_{t1_{w_2}})\\
        ...\\ 
        RelTypeSim(\vec{r}_{tn_{w_1}},\vec{r}_{tn_{w_2}})
    \end{matrix}\right\}
\end{equation}
Finally, the overall relational similarity between two terms is defined as the normalized sum of the weighted type-wise relational similarities. The normalization is based on the total prevalence of all common relation types, as shown in the following equation:
\begin{equation}\label{eq:S3RelSim}
    RelSim(w_1,w_2)=\frac{1}{\sum\limits_{r_t}{P(r_t)}}\times\sum\limits_{r_t}{P(r_t)\times RelTypeSim(r_{t_{w_1}},r_{t_{w_2}})},
\end{equation}
where $r_t\in RTSet(w_1)\cap RTSet(w_2)$.

\subsection{Relatedness} \label{subsec:Relatedness}
Relatedness can be expressed by the direct connection(s) between two terms through non-taxonomic relations. It is a reflection of the contextual bond between terms. As described in the gold standard datasets, MC \cite{mc-28-1991}, RG \cite{rg-65-1965}, WordSim \cite{wordsim-353-2001}, and MTurk \cite{halawi2012MTurk771}, concepts were evaluated based on their similarity and relatedness. 
For example, in \cite{wordsim-353-2001}, one instruction was "\textit{When estimating similarity of antonyms, consider them "similar" (i.e., belonging to the same domain or representing features of the same concept), rather than "dissimilar"}".
Therefore, it is essential to incorporate relatedness into a similarity measure, as shown in Eq. \eqref{eq:poly-rel}. 

A non-taxonomic path associates two terms with a relationship that is reflected by the intermediate relations between them. Intuitively, the longer the path between terms, the less related they are. Furthermore, each relation has a weight that reflects its importance to the domain. Hence, relations with higher weights indicate a stronger contextual relationship between the associated terms. Inversely, the weaker the weight between terms, the less related they are. Finally, terms that have multiple paths indicate stronger bond between them. For instance, the lexical terms (king and queen) share more than one path between them, as they are considered antonyms in addition to their respective synsets, which are member-meronyms of the synset "royal family". Based on our results, these terms would have a relatedness of 92\% over both paths. Based on these principles, we propose a new relatedness measure based on weighted non-taxonomic relational paths. 

\subsubsection*{Strategy 4}\label{subsubsec:Strategy 4}
\begin{figure}[t]
    \begin{subfigure}[b]{\textwidth}
        \includegraphics[width=.9\textwidth]{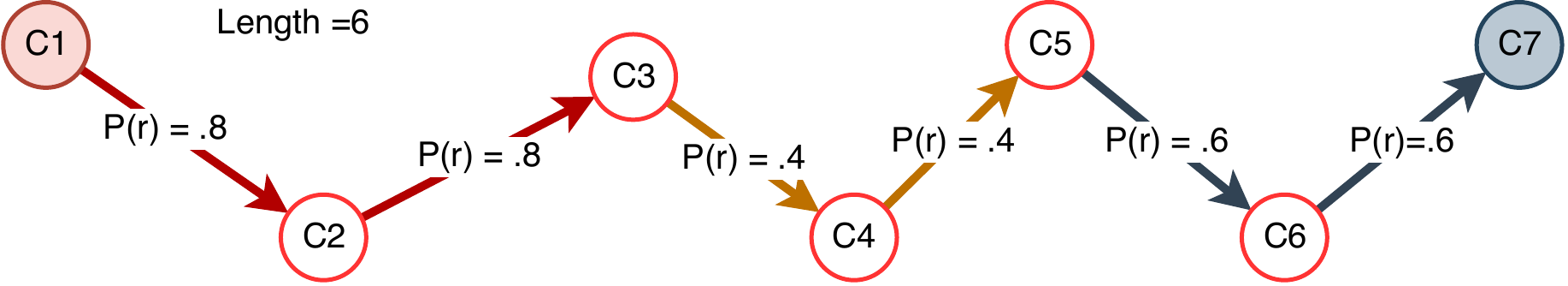}
        \captionsetup{font=small,labelfont=small}
        \caption{1-Distance(C1,C7) = 0.833}
        \label{fig.sub:Rel:a}
        \vspace{.5cm}
    \end{subfigure}\\
    \begin{subfigure}[b]{\textwidth}
        \includegraphics[width=.9\textwidth]{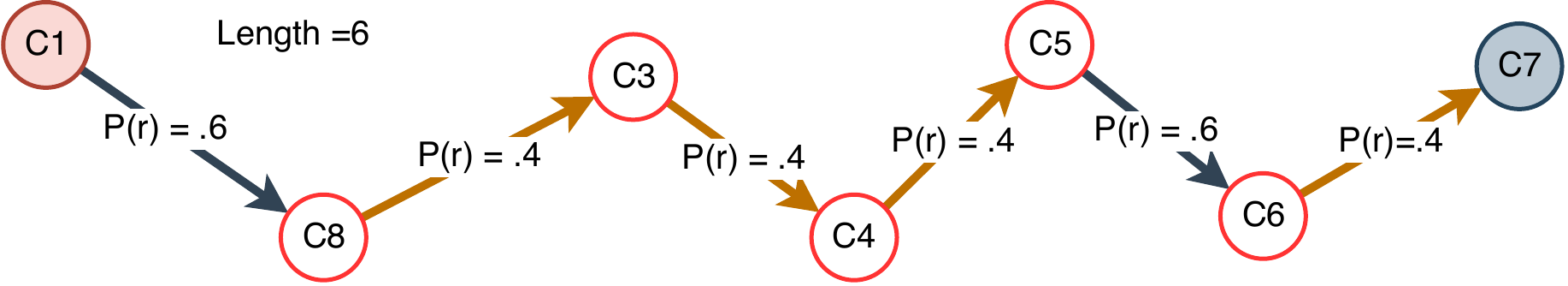}
        \captionsetup{font=small,labelfont=small}
        \caption{1-Distance(C1,C7) = 0.811}
        \label{fig.sub:Rel:b}
        \vspace{.5cm}
    \end{subfigure}\\
    \begin{subfigure}[b]{\textwidth}
        \includegraphics[width=.6\textwidth]{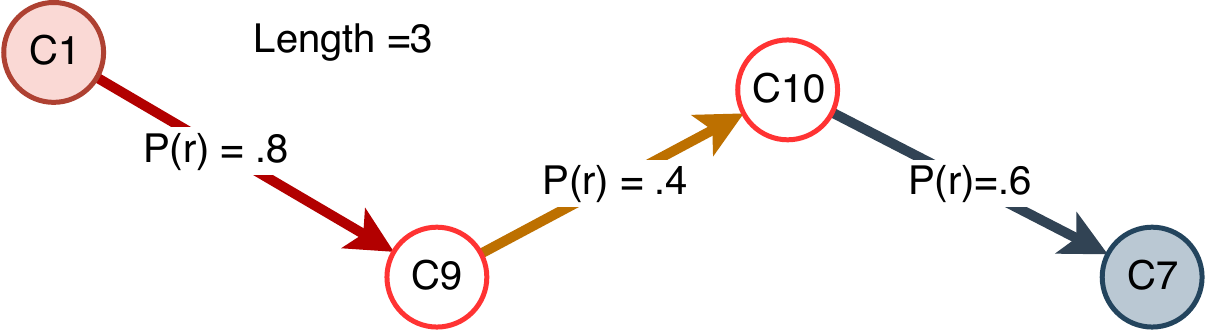}
        \captionsetup{font=small,labelfont=small}
        \caption{1-Distance(C1,C7) = 0.916}
        \label{fig.sub:Rel:c}
    \end{subfigure}
    \caption{Relatedness paths between two concepts with different relations prevalence }
    \label{fig:S4_RelatednessPaths}
\end{figure}

Taking the above considerations into account, we propose a relatedness measure as a function of the number of paths, their length and strength. The length of a path is a function of the number of non-taxonomic relations between terms. Strength is measured by the proposed prevalence of the path relations as defined in Section \ref{subsec:Relations prevalence}. As the path length increases, the relatedness decreases. On the other hand, as the accumulated weight of all relations in the path increases, relatedness increases. Therefore, relatedness needs to be a monotonically decreasing function with respect to path length, while monotonically increasing with respect to the relations' weight. To satisfy these constraints, we propose the following relatedness measure:
\begin{equation}\label{eq:S4Relatedness}
    Relatedness(w_1, w_2)=\frac{1}{n}\times\sum_{path_i=1}^{n}{1-Distance_i(w_1, w_2)},
\end{equation}
where $n$ is the number of paths, and the distance is calculated as shown below:
\begin{equation}\label{eq:S4DistanceRIC}
    Distance_i(w_1, w_2) = \frac{1}{max\_depth_{wn}}\times\sum\limits_{r_t\in path_i(w_1,w_2)}{e^{-P(r_t)}},
\end{equation}
where $max\_depth_{wn}$ is the the maximum hierarchical depth of WordNet. To ensure paths convey meaningful relatedness between terms, we only consider paths with a length shorter or equal to $max\_depth_{wn}$. Fig. \ref{fig:S4_RelatednessPaths} illustrates the effects of path-length and relation-prevalence on the relatedness measure. As can be seen from Fig. \ref{fig.sub:Rel:a} and Fig. \ref{fig.sub:Rel:b}, where the pathways between concepts C1 and C7 have the same length but different prevelances,  the higher the prevalences between concepts the greater the relatedness. However, for Fig. \ref{fig.sub:Rel:c}, although the pathway between C1 and C7 has the same average prevalence per relation type as that in Fig. \ref{fig.sub:Rel:a}, yet its relatedness is higher due to the shorter path. Finally, the overall relatedness measure between concepts C1 and C7 over all of the three paths is 0.854, which represents the average relatedness over all of the paths as shown in Eq. \ref{eq:S4Relatedness}.
\section{Evaluation and experimental results}\label{sec:Evaluation and Experimental Results}
The goal of our experiments is to evaluate the proposed semantic similarity and relatedness based on the WordNet database and the most commonly-used gold standard datasets. This section presents WordNet KG, datasets, evaluation metrics, implementation, and detailed discussion of the results.

\subsection{WordNet}\label{subsec:WordNet}
WordNet is an English-based lexical database, where words are organized within each Part of Speech (POS) category into sets of cognitive synonyms named synset. Synsets are organized into a hierarchical structure (taxonomy) from the most abstract concepts (i.e., entity in the nouns category) to the most specific leaf concepts. Synsets and lexical terms are linked through means of pointers that represent a specific semantic relationship between them. 

The main and most structured explicit relation type is the hyponymy (ISA) relation and its inverse hypernymy relation, which forms the hierarchical structure of WordNet. ISA represents the generalization/specialization relation between concepts. The next most common relation used is holonomy/meronymy (part of) relation. All other relations such as antonymy, theme and derivation are used much less frequently. Tables \ref{tb:wn_RelationLinksSyn} and \ref{tb:wn_TaxonLinks} depict the frequency of each relation used within WordNet\footnote{The frequency is calculated based on our implementation focusing on the nouns category from WordNet 3.1}. 

WordNet was designed with the intention of providing a machine-readable dictionary that determines a word definition through semantic relations \cite{miller1995wordnet}. In addition to the explicit relations, lexical terms within one synset also share implicit relations between each other, namely “synonymy”. As shown in Table \ref{tb:wn_RelationLinksSyn}, this relation is explicitly emphasized in our approach. We use the synset\_member relation that exists in WordNet to identify the synonyms of each synset. The number of synonym relations is then computed using the combination $\binom{n}{2}$ where \textit{n} is the number of synonyms (synset\_member) within one synset. Hence, the use of synonymy relation comply with the spirit of WordNet, providing semantic definitions for its synsets and lexical terms. Therefore, in this work, we employ both of explicit and implicit semantic relations to obtain better semantic similarity and relatedness between terms.

\begin{table}
\centering
	\caption{Hierarchical relations in WordNet}
	\label{tb:wn_TaxonLinks}
		\begin{tabular}{p{.35\columnwidth}>{\raggedleft}m{.25\columnwidth}r}
			\hline
			\textbf{Relation Name} & \textbf{Frequency} & \textbf{Prevalence} \\
			\hline 
			\textbf{hyponym} 			& 75180 & 44.89 \%		\\
			\textbf{hypernym} 			& 75139 & 44.86 \% 		\\
			\textbf{instance\_hyponym}	& 8592 	& 5.13 \% 		\\
			\textbf{instance\_hypernym} & 8568 	& 5.12 \% 		\\
			\hline
		\end{tabular}
\end{table}

\subsection{Gold standard datasets}\label{subsec:Gold Standard Datasets}
Our experimental environment consists of the lexical database WordNet 3.1 as a KG and four widely-used evaluation datasets. The first dataset is a study done by Rubenstein and Goodenough\cite{rg-65-1965}. It contains 65 pairs of words (RG), where each pair is assigned a rating of similarity ranging between [0,4]; 0 being most dissimilar/unrelated terms, and 4 is exactly similar. The second dataset was compiled by Miller and Charles \cite{mc-28-1991}. It contains 28 pairs of words (MC), chosen carefully from RG to represent a full range of similarity/relatedness. The third dataset was conducted by Finkelstein et al. \cite{wordsim-353-2001}, and contains 353 pairs of terms (WordSim). The rating range is between [0,10], where 0 is completely dissimilar/unrelated and 10 is exactly similar. The instructions provided to the subjects --- humans experts --- were to assign a rating for each pair based on the similarity/relatedness of the two words. More specifically, in WordSim, the instructions were clear for the antonym words to be considered as similar rather than dissimilar, since they belonged to the same domain and/or express feature of the same topic. Finally, a more comprehensive larger dataset, MTurk, is used to demonstrate robustness and scalability of the proposed approach. MTurk is a gold standard compiled by Halawi and Dror on Amazon Mechanical Turk (MTurk), intended to emphasize relatedness between terms. MTurk contains 771 pairs or words with their average relatedness score between [1-5], where 1 represents "not related" words, and 5 represents "highly related" words \cite{halawi2012MTurk771}. It is worth mentioning that there exists other large datasets that we didn't use in our experiment as they explicitly quantify similarity measure without relatedness while our approach is more general and considers both similarity and relatedness (i.e. SimLex-999) \cite{hill2015simlex}. 

Table \ref{tb:GoldStandard_Characteristics} provides a summary of the main characteristics and statistics of each dataset. In our implementation, we examine all pairs within each dataset, with the exception of WordSim as it contains few pairs that were not found in WordNet-3.1 due to variation in the term morphology or tense. The last four rows in the table (rows 5, 6, 7, and 8) are the focus of our poly-relational approach. Row 5 shows the number of pairs that share one or more non-taxonomic relation type, regardless of whether or not the object of the relation is the same. For instance, "car has \textit{part-meronym} ..." and "auto has \textit{part-meronym}...", hence both terms share the same relation type \textit{part-meronym}. Row 6 counts the number of terms where one or more of its non-taxonomic relations contain multiple instances. For example, for "car has \textit{part-meronym} fender, car has \textit{part-meronym} engine", there are two instances of the relation \textit{part-meronym}. Row 7 shows the number of pairs where at least one of its terms matches a term contained in row 6. Finally, row 8 describes the number of pairs that have at least one non-taxonomic path that connects the pair (i.e., "Car is \textit{synonym} of auto", "King is member-meronym of royal family, and royal family is member-holonym of queen". 

\begin{table}
\centering
	\caption{Gold standard datasets characteristics}
	\label{tb:GoldStandard_Characteristics}
	\resizebox{\textwidth}{!}{
		\begin{tabular}{@{\makebox[.1\columnwidth][c]{\rownumber}}lcccc}
			\hline
			\textbf{Criteria} & \textbf{MC} & \textbf{RG} & \textbf{WordSim} & \textbf{MTurk} 
			\gdef\rownumber{\stepcounter{magicrownumbers}\arabic{magicrownumbers}} \\
			\hline
			\textbf{\# of pairs} 						& 28 & 65 & 353 & 771 \\
			\textbf{\# of pairs (our implementation)} 	& 28 & 65 &	342	& 770 \\
			\textbf{Distinct terms/senses}				& 44 & 65 & 500	& 1281 \\
			\textbf{Terms with non-taxo. relations} 	& 18 & 23 &	207	& 487 \\
			\textbf{Pairs with common relations (CR)} 	& 4  & 4  & 54 & 110 \\ 
			\textbf{Terms with multi-instances (MI)} 	& 8  & 10 & 77 & 204 \\ 
			\textbf{Pairs with CR \& MI} 	            & 2  & 2 & 20 & 48	\\ 
			\textbf{Pairs with one or more path(s)} 	& 6  & 12 & 35 & 149 \\
			\hline
		\end{tabular}
	}
\end{table}

\subsection{Evaluation metrics}\label{subsec:Evaluation Metrics}
The practice of evaluating semantic similarity measures has relied on the correlation between the proposed method and gold standard \cite{seco2004,zhou2008IC,sebti2008,sanchez2011IC,cai2018hybrid,meng2012,zhang2018}. Two main correlation methods have been applied. The first is the Pearson correlation for two random variable X and Y, as shown below:
\begin{equation}\label{eq:CorrPearsonRho}
    \rho = \frac{\text{cov}(X,Y)}{\sigma_x \sigma_y}
\end{equation}

Yet, a simplified estimated correlation of the Pearson has been normally applied based on the following:
\begin{equation}\label{eq:CorrPearsonEst}
    r = \frac{{}\sum\limits_{i=1}^{n} (x_i - \overline{x})(y_i - \overline{y})}
{\sqrt{\sum\limits_{i=1}^{n} (x_i - \overline{x})^2}\sqrt{\sum\limits_{i=1}^{n} (y_i - \overline{y})^2}},
\end{equation}
where $n$ is the size of the sampled sets, $x_i,y_i$ are the $i^{th}$ element of semantic similarities reported by any measure, and the human judgment, respectively, and $\overline{x}, \overline{y}$, represent the mean for each set.

The second correlation method is the Spearman correlation coefficient (Spearman's rho). This is a rank-based correlation, which is not restricted to continuous data.
\begin{equation}\label{eq:CorrSpearman}
    \rho = 1- {\frac {6 \sum d_i^2}{n(n^2 - 1)}},
\end{equation}
where $d$ is the pairwise distance of the ranks of the elements $x_i$ and $y_i$, and $n$ is the size of the sampled set.

\subsection{Implementation}\label{subsec:Experiments}
Our experimental implementation utilizes the .NET framework and dotNetRDF Library\footnote{\url{http://www.dotnetrdf.org/api/html/N_VDS_RDF.htm}}. This library helps to convert the N-Triples of WordNet 3.1\footnote{\url{http://wordnet-rdf.princeton.edu/}} KG file into an in-memory Resource Description Framework graph (RDF graph). Out of the 5.5 million triples, we focus only on noun synsets and lexical senses, including their relations. However, some of these relations, such as translation, gloss, and tag counts, are also ignored because they serve various purposes that are outside the scope of this research. In summary, we focus on triples with relations that serve the semantic of the noun synsets and lexical senses. These relations are listed in Tables \ref{tb:wn_RelationLinksSyn} and \ref{tb:wn_TaxonLinks}. After the extraction and cleansing, we are left with a total of 81,816 noun synset, 262,786 individual lexical senses, and 374,453 instances relations. These relations are grouped into 12 non-taxonomic and four taxonomic relation types as shown in Tables \ref{tb:wn_RelationLinksSyn} and \ref{tb:wn_TaxonLinks}.

It is worth mentioning that the ontology of WordNet\footnote{\url{http://wordnet-rdf.princeton.edu/ontology}} is very abstract, and the relations have very shallow hierarchy with no domain and range concepts. Hence, in our experiment, the taxonomic IC ($IC_{Tax}(r_t)$), and the global IC ($IC_{GC}(r_t)$) do not contribute towards calculating the relation IC ($RIC(r_t)$). As a result, relations attain their information strictly from the local IC ($IC_{LC}(r_t)$) from WordNet's KG. 

The rest of the implementation consists of an offline module that calculates WordNet's KG statistics, and the gold standard datasets statistics. In addition, we implemented the existing baselines' IC and taxonomic similarity measures.

\subsection{Results analysis and discussion}\label{subsec:Results and Analysis}
The main purpose of this experiment is to confirm the observations in section \ref{subsec:Motivations and Contribution}, which emphasize that all relations convey an important informative component of a concept's semantic definition and information content, and that relation type prevalence reflects its relevance to the modeled domain. This is demonstrated in this section by showing the enhancements provided by our poly-relational approach to existing taxonomic-based similarity measures. The common practice of evaluating any approach is to compute the pairwise similarities of a given dataset. Then, calculate the correlation as discussed in section \ref{subsec:Evaluation Metrics}, with the gold standard for that dataset. The higher the correlation with the gold standard, the better the approach. Therefore, for each baseline, we implement its IC metric to evaluate non-taxonomic relations. Then employ Eq. \eqref{eq:poly-rel_S1S2S3} to incorporate the proposed $RelSim(w_1,w_2)$. Finally, compare our results with each baseline based on gold standard correlations.

In section \ref{sec:Proposed method} we presented four strategies, which will be referred to in this Section as S1, S2, S3, and S4. The results of each strategy, as shown in Fig. \ref{fig:S1Results} to Fig. \ref{fig:S4Results}, are described by the gain between the proposed approach and each baseline. The gain is based on the correlation of each approach with the gold standard, which is computed as follows: 
\begin{equation}\label{eq:gain}
    gain= \frac{Corr()_{PolyR}-Corr()_{baseline}}{Corr()_{baseline}},
\end{equation}
where $Corr()_{PolyR}$ and $Corr()_{baseline}$ are the correlations of the poly-relational approach and any given baseline approach with the gold standard dataset, respectively. Each figure
shows the gain for its respective strategy, and includes the evaluation using the Pearson correlation, as defined in Eq. \eqref{eq:CorrPearsonEst}, and the Spearman correlation, as defined in Eq. \eqref{eq:CorrSpearman}.

Finally, we compute the full semantic similarity and relatedness as described in Eq. \eqref{eq:poly-rel}. However, since S1, S2, and S3 do not involve relatedness, semantic similarity (\textit{SemSim}) for these strategies includes only the taxonomic similarity and the non-taxonomic relational similarity measures, as shown in the equation below:
\begin{equation}\label{eq:poly-rel_S1S2S3}
    SemSim(w_1,w_2)= (1-\alpha_1) \times TaxSim(w_1,w_2)+ \alpha_1 \times RelSim(w_1,w_2),
\end{equation}
where $\alpha_1$ is the contribution of the relational similarity that overcomes the limitations of the existing baseline taxonomic similarity. This is measured by the prevalence of the non-taxonomic relations common to both terms, as in the following equation:
\begin{equation}\label{eq:S1-3alph}
    \alpha_1 = \sum\limits_{r_t\in R(w_1) \cap R(w_2)} {P(r_t)}
\end{equation}
Strategy 4, on the other hand, incorporates the relatedness in the overall semantic similarity and relatedness measure.

\subsubsection*{Strategy 1}\label{subsubsec:S1Results}
\begin{figure}
	\centering
	\includegraphics[trim=0 0 0 45,clip,width=\linewidth]{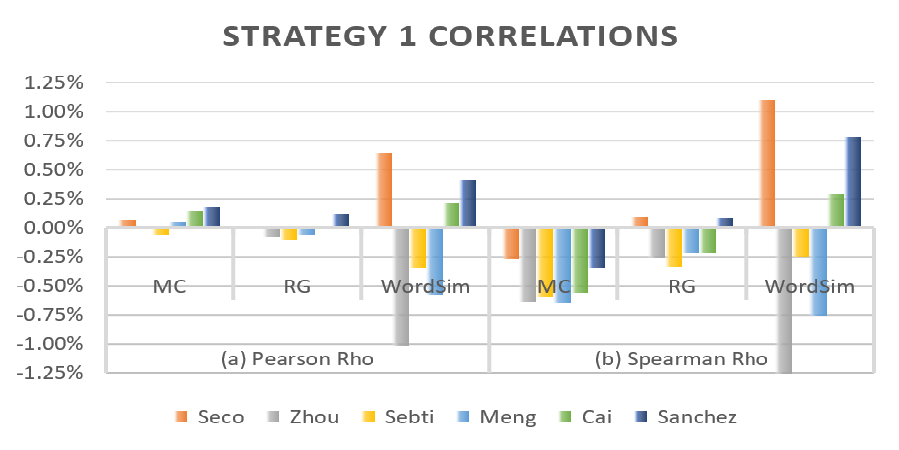}
	\caption{Semantic similarity gain using strategy 1}
	\label{fig:S1Results}
\end{figure}

As stated in Section \ref{subsubsec:Strategy 1}, S1 considers all non-taxonomic relations and encapsulates their semantic IC values into a single attribute for each term, namely $SemIC$. The results for this strategy, as presented in Fig. \ref{fig:S1Results}, demonstrates limited improvements across the evaluated baselines and gold standard datasets. For instance, using MC gold standard, although our approach demonstrates a gain of $0.05\%$ with \textit{Meng} as a baseline, it shows a consistent decline with the two other datasets, especially when using the Spearman correlation. Similarly, with MC gold standard using \textit{Seco} as a baseline, the gain is approximately $0.06\%$, while it shows a decline of $0.27\%$ with the Spearman correlation. Examining the same baseline with WordSim gold standard dataset, our approach has approximately $0.64\%$ and $1.10\%$ gain with both the Pearson and the Spearman correlations, respectively. 

The decline in this strategy is due to the fact that existing taxonomic IC-based similarity measures rely on the IC of the LCS of the two terms, as shown in in Table \ref{tb:SimMetric}. However, non-taxonomic relations do not follow hierarchical structure, and therefore, traditional taxonomic IC-based similarity measures are not effective for evaluating semantic IC. This was the main motivation for investigating new ways for computing relational-based similarity in subsequent strategies. Furthermore, S1 does not take into consideration the type of non-taxonomic relations when comparing the semantic IC values of two terms. Thus, ignoring an important aspect of terms' semantic.

\subsubsection*{Strategy 2}\label{subsubsec:S2Results}
Strategy 1 has clearly undermined baselines similarities instead of improving them. As discussed in Section \ref{subsubsec:Strategy 2}, S2 overcomes the limitations of S1 by expanding the encapsulated $SemIC$ of a term. Instead of reducing $SemIC$ to a single-value attribute, terms are attributed a vector of relation type-based $RIC$ measures, each of which represents a single relation type (part-meronym, instance-meronymy, derivation, etc.), as shown in Eq. \eqref{eq:S2_SemIC}. 

\begin{figure}
	\centering
	\includegraphics[trim=1 0 0 45,clip,width=\linewidth]{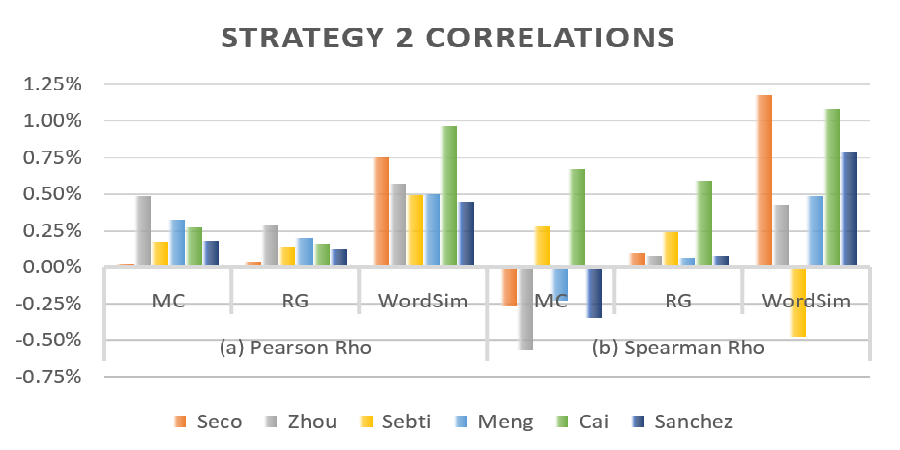}%
	\caption{Semantic similarity gain using strategy 2}
	\label{fig:S2Results}
\end{figure}

As shown in Fig. \ref{fig:S2Results}a, with the exception of MTurk dataset using \textit{Seco, Zhou, Meng}, and \textit{s\'anchez}, S2 demonstrates a consistent gain across all baselines and gold standard datasets. For instance, using WordSim gold standard dataset and \textit{Meng} as a baseline, S1 shows a decline of $-0.58\%$, while S2 shows a gain of $0.50\%$. This remains consistent with the Pearson correlation using the other datasets too. However, based on the Spearman correlation, as shown in Fig. \ref{fig:S2Results}b, although there is an overall improvement with RG and WordSim gold standard datasets, MC still demonstrates a decline with most baselines. We believe this is due to the sensitivity of the Spearman ranking correlation on such a small dataset, as the same baselines, with a relatively larger dataset (RG), still demonstrate a minimal gain.

Despite the decline in MC dataset using the Spearman correlation, improvements in all other evaluations motivated us to go beyond and expand each vector element of $SemIC$ into existing instance relations of that type. This is the main motivation behind S3. 

\subsubsection*{Strategy 3}\label{subsubsec:S3Results}
\begin{figure}[!t]
	\centering
	\includegraphics[trim=0 0 0 45,clip,width=\linewidth]{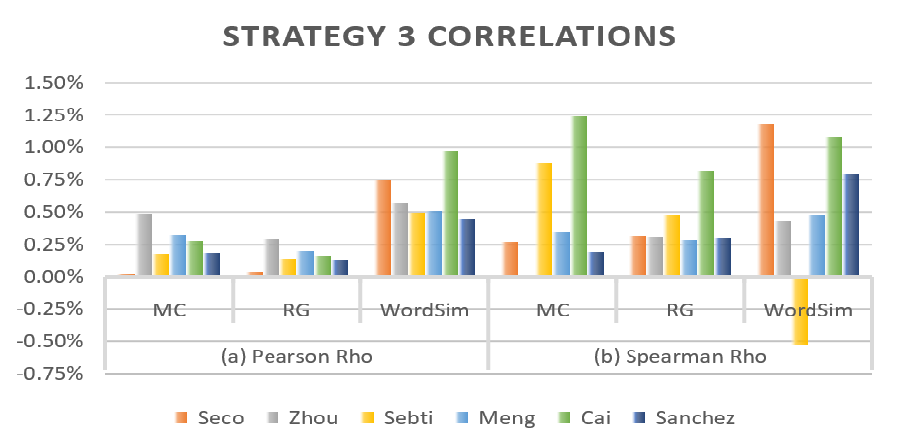}
	\caption{Semantic similarity gain using strategy 3}
	\label{fig:S3Results}
\end{figure}

Fig. \ref{fig:S3Results}a shows the gain of S3 using the Pearson correlation. It can be observed that the gain nearly mirrors S2 results. This can be explained by the fact that only a small proportion of the used datasets exhibit pairs with common relations that have multiple instances, as shown in Table \ref{tb:GoldStandard_Characteristics}. Therefore, their effect is not significant on the overall results. On the other hand, using the Spearman correlation, a significant improvement can be observed for MC and RG datasets as shown in Fig. \ref{fig:S3Results}b. This can be justified by the fact that the Spearman correlation, being based on ranking, is very sensitive to the size of the used dataset. For a small dataset, semantic similarity changes to few pairs will significantly impact the overall ranking, thus resulting into considerable improvement in the correlation. Inversely, for a large dataset, semantic similarity changes to few pairs will not have the same impact on the overall ranking, thus resulting in minor improvement of the correlation. This is clearly shown in Fig. \ref{fig:S3Results}b for MC and RG, which are small datasets with only 2 pairs affected by semantic similarity changes. However, for WordSim dataset, which includes 342 pairs, and only 20 pairs affected by semantic similarity change, the improvement in correlation is very minor.

\begin{figure}[!ht]
	\centering
	\includegraphics[trim=2 0 0 75,clip,width=\linewidth]{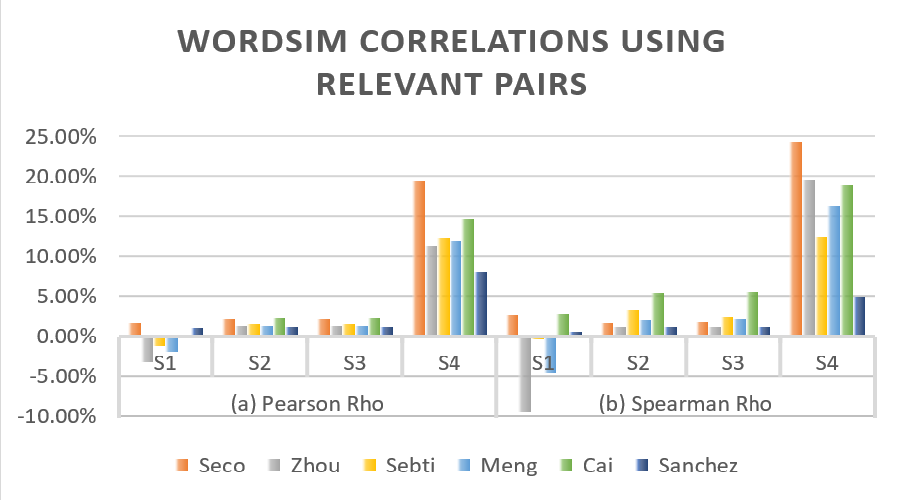}
	\caption{Semantic similarity gain for all strategies using WordSim relevant pairs}
	\label{fig:S1_4WordSimReleventPairs}
\end{figure}

The only surprising mystery remains with a decline using the \textit{Sebti} baseline with WordSim dataset. After analyzing the baseline, dataset, and other literature \cite{zhu2016Computing}, we conclude that this could be caused by a small subset of the dataset that is affected by \textit{Sebti's} calculation method. This finding is confirmed in another literature \cite{zhu2016Computing}, noticing similar behaviour of abnormality with the same dataset. To address the shortcomings in WordSim dataset, we conducted another experiment using only a subset of the whole dataset, focusing mainly on relevant pairs. These are pairs with at least one common relation type, or at least one connected path, or both. In particular, $54$ pairs were used to test strategies S1, S2, and S3, and $66$ pairs were used to test S4. The use of this semantically rich dataset shows consistent gain across S2, S3, and S4 as can be observed in Fig. \ref{fig:S1_4WordSimReleventPairs}. On the other hand, the inconsistent gain in S1 confirms our initial observation for S1, that semantic-blind comparison is not effective.
\subsubsection*{Strategy 4}\label{subsubsec:S4}
\begin{figure}[!b]
	\centering
	\includegraphics[trim=0 0 0 45,clip,width=\linewidth]{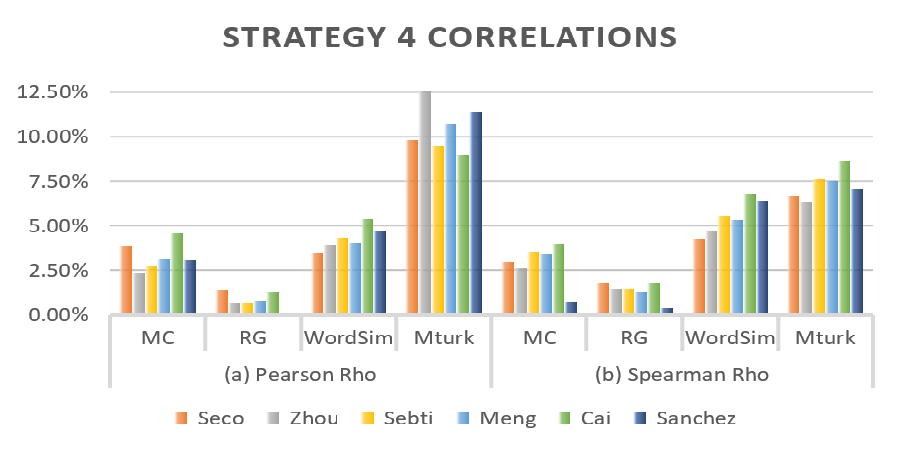}
	\caption{Semantic similarity and relatedness gain using strategy 4}
	\label{fig:S4Results}
\end{figure}

As described in Section \ref{subsubsec:Strategy 3}, S4 differs from the previous strategies by incorporating a relatedness measure, which is based on weighted paths between terms, as defined in Eq. \eqref{eq:S4Relatedness}. The overall semantic similarity and relatedness measure that was defined in Eq. \eqref{eq:poly-rel}, is expressed as follows:
\begin{align}\label{eq:poly-rel_S4}
    SemSimRel&(w_1,w_2) = (1-\alpha_2-\beta) \times TaxSim(w_1,w_2) + \nonumber \\ &\alpha_2 \times RelSim(w_1,w_2) +\beta \times Relatedness(w_1,w_2),
\end{align}
where $\alpha_2$ and $\beta$ measure the contribution of relational similarity and relatedness, respectively. Note that $\alpha_2$ in Eq. \eqref{eq:poly-rel_S4} is different from $\alpha_1$ in Eq. \eqref{eq:poly-rel_S1S2S3}, due to the effect of the relatedness parameter in S4. The optimal values for $\alpha_2$ and $\beta$ in S4 are empirically evaluated. We used different combinations of ($\alpha_2,\beta$), and we found that the optimal values across all datasets and baselines are $\alpha_2=0.12$ and $\beta=0.55$.

To test strategy 4 with a more relevant dataset that includes both similarity and relatedness, we used the MTurk gold standard dataset. As described in section \ref{subsec:Gold Standard Datasets}, MTurk was built to capture relatedness measure between terms, which is the main focus of strategy 4. Based on the results shown in Fig. \ref{fig:S4Results}, the gains for this strategy are consistent across all baselines and gold standard datasets. For example, using \textit{Sebti} as a baseline, S3 shows a decline with the Spearman correlation, while S4 shows over $5\%$ of gain. Also, the results show better performance with MTurk, which is a larger relatedness dataset, thus confirming scalability of S4. 
As shown in Fig. \ref{fig:S4Results}, the highest gain is $12.63\%$, which is achieved based on \textit{Zhou} baseline with MTurk gold standard dataset using Pearson's correlation. On the other hand, the highest correlation value for the same dataset is attained using \textit{Cai} baseline with the pearson correlation of 0.8620, see Table \ref{tb:PearsonCorrResults}.
As shown in Fig. \ref{fig:S4Results}, the proposed poly-relational approach demonstrates consistent improvement in the semantic similarity and relatedness measure across all datasets.
These results are coherent with the human perception of semantic similarity and relatedness within the gold standard datasets. 

\begin{table}[!t]
\renewcommand{\tabcolsep}{0.1cm}
	\caption{Pearson correlation with gold standard and proposed strategies}
	\label{tb:PearsonCorrResults}
	\resizebox{\columnwidth}{!}{
		\begin{tabular}{lllll}
			\hline
			\textbf{Method} & \textbf{MC} & \textbf{RG} & \textbf{WordSim} & \textbf{MTurk} \\
			\hline  
			\textit{RES}\textsubscript{corpus}&	0.8154	&	0.8475	&	0.5370	&	0.7240 	\\
			\textit{WUP}\textsubscript{corpus}&	0.7740	&	0.8047	&	0.4256	&	0.6735	\\
			\textit{LIN}\textsubscript{corpus}&	0.7759	&	0.7569	&	0.4376	&	0.6151	\\
			\textit{PATH}\textsubscript{is-a}&	0.7504	&	0.7889	&	0.4295	&	0.6732	\\
			\textit{Seco}\textsubscript{graph}&	0.8943	&	0.8680	&	0.3528	&	0.4534	\\
			\textit{Zhou}\textsubscript{graph}&	0.8429	&	0.8403	&	0.3259	&	0.4414	\\
			\textit{Sebti}\textsubscript{graph}&	0.8434	&	0.8577	&	0.3835	&	0.4564	\\
			\textit{Meng}\textsubscript{graph}&	0.8563	&	0.8711	&	0.3652	&	0.4702	\\
			\textit{Cai}\textsubscript{graph}   &	0.8635	&	0.8840	&	0.3856	&	0.4898	\\
			\textit{s\'anchez}\textsubscript{graph}&	0.8950	&	0.8701	&	0.3475	&	0.4489	\\
			\textit{TransE}\textsubscript{corpus}&	0.8310	&	0.7737	&	0.4032	&	0.4465	\\
			\textit{TransH}\textsubscript{corpus}&	0.7800	&	0.8041	&	0.4097	&	0.4289	\\
			\textit{TransG}\textsubscript{corpus}&	0.8102	&	0.7720	&	0.3639	&	0.4083	\\
			\textit{Strategy 1}\textsubscript{graph} (baseline) &	0.8966	(s\'anchez ) &	0.8841	(Cai) &	0.3864	(Cai) &	0.4902	(Cai) \\
			\textit{Strategy 2}\textsubscript{graph} (baseline)	&	0.8966	(s\'anchez ) &	0.8854	(Cai) &	0.3894	(Cai) &	0.4900	(Cai) \\
			\textit{Strategy 3}\textsubscript{graph} (baseline)	&	0.8966	(s\'anchez ) &	0.8854	(Cai) &	0.3894	(Cai) &	0.4900	(Cai) \\
			\textit{Strategy 4}\textsubscript{graph} (baseline)	&\textbf{0.9291} (Seco) &\textbf{0.8953} (Cai) &\textbf{0.7079} (Seco) &\textbf{0.8620} (Cai) \\
			\hline
		\end{tabular}
	}
\end{table}
\begin{table}[!t]
\renewcommand{\tabcolsep}{0.05cm}
	\caption{Spearman correlation with gold standard and proposed strategies}
	\label{tb:SpearmanCorrResults}
	\resizebox{\columnwidth}{!}{
		\begin{tabular}{lllll}
			\hline
			\textbf{Calculation Method} & \textbf{MC} & \textbf{RG} & \textbf{WordSim} & \textbf{MTurk} \\
			\hline  
			\textit{RES}\textsubscript{corpus}	&	0.8028	&	0.8067	&	0.6082	&	0.7020	\\
			\textit{WUP}\textsubscript{corpus}	&	0.7681	&	0.7766	&	0.4278	&	0.7224	\\
			\textit{LIN}\textsubscript{corpus}	&	0.7457	&	0.6952	&	0.4751	&	0.6758	\\
			\textit{PATH}\textsubscript{is-a}	   &	0.7506	&	0.8002	&	0.3869	&	0.7271	\\
			\textit{Seco}\textsubscript{graph}	&	0.8660	&	0.7963	&	0.3308	&	0.4896	\\
			\textit{Zhou}\textsubscript{graph}	&	0.8130	&	0.7883	&	0.3232	&	0.5057	\\
			\textit{Sebti}\textsubscript{graph}	&	0.7781	&	0.7693	&	0.3349	&	0.4762	\\
			\textit{Meng}\textsubscript{graph}	&	0.8001	&	0.7918	&	0.3205	&	0.5082	\\
			\textit{Cai}\textsubscript{graph}	    &	0.8174	&	0.7981	&	0.3150	&	0.5025	\\
			\textit{s\'anchez}\textsubscript{graph}	&	0.8772	&	0.7994	&	0.3195	&	0.5036	\\
			\textit{Zhang}\textsubscript{graph}	    &	0.3480	&	0.3434	&	0.1013	&	0.2195	\\
			\textit{TransE}\textsubscript{corpus}	&	0.8085	&	0.7236	&	0.3525	&	0.4622	\\
			\textit{TransH}\textsubscript{corpus}	&	0.7670	&	0.7406	&	0.3728	&	0.4454	\\
			\textit{TransG}\textsubscript{corpus}	&	0.8342	&	0.6689	&	0.3180	&	0.4203	\\
			\textit{Strategy 1}\textsubscript{graph} (baseline)	&	0.8741	(s\'anchez ) &	0.8001	(s\'anchez ) &	0.3344	(Seco ) &	0.5035	(s\'anchez) \\
			\textit{Strategy 2}\textsubscript{graph} (baseline)	&	0.8741	(s\'anchez ) &	0.8028	(Cai) &	0.3347	(Seco ) &	0.5029	(s\'anchez ) \\
			\textit{Strategy 3}\textsubscript{graph} (baseline)	&	0.8788	(s\'anchez ) &	0.8046	(Cai) &	0.3347	(Cai) &	0.5039	(s\'anchez ) \\
			\textit{Strategy 4}\textsubscript{graph} (baseline)	&\textbf{0.8917} (Seco) &\textbf{0.8121} (Cai) &\textbf{0.6848} (Seco) &\textbf{0.8600} (Cai) \\
			\hline
		\end{tabular}
	}
\end{table}
To test the robustness of the proposed approach, we further extended our experiment to include WordNet's existing similarity measures \cite{pedersen2004wordnet}, which includes some corpus-based measures
, such as RES
, WUP
, and LIN
, as well as taxonomic-based path measure PATH 
\cite{pedersen2004wordnet}
. Furthermore, we compared our approach with the state of the art Knowledge Graph Embedding semantic similarity models implemented in \cite{yu2019pykg2vec}. We used the implementation provided in \cite{yu2019pykg2vec} to train three models (TransE, TransH, and TransG) on WordNet and the gold standard datasets. We have then  obtained the embedding vectors for each term and computed the cosine similarity for each pair of terms. Tables \ref{tb:PearsonCorrResults} and \ref{tb:SpearmanCorrResults} display the actual Pearson and Spearman correlations based on our implementation using WordNet 3.0, the KGE from \cite{yu2019pykg2vec} and the NLTK WordNet similarity implementations. Also, the tables present the respective correlations for the six examined benchmarks \cite{seco2004,zhou2008IC,sebti2008,meng2012,cai2018hybrid,sanchez2011IC}, in addition to the poly-relational approach, showing the best obtained correlation across all baselines for each strategy. It can be seen from the results provided in Tables \ref{tb:PearsonCorrResults} and \ref{tb:SpearmanCorrResults} that the proposed strategies outperform all baselines. Furthermore, the results show gradual improvement from S1 to S4, with the exception of S3 when using the Pearson correlation. This is justified above, while discussing the results of S3. As highlighted in both tables, the proposed method provides significant improvement over all baselines with both the Pearson and the Spearman correlations, thus showing its superiority over all other methods.

Finally, we analyzed the complexity of the proposed semantic similarity and relatedness algorithm in terms of the used IC function for each baseline, and concepts' depth in the WordNet graph. It should be noted here that the number of relation types in the WordNet graph is constant, and therefore it does not affect the computational complexity of the proposed technique. As shown in Eq.\ref{eq:poly-rel}, the complexity of the proposed method is given by the maximum complexity among the three algorithms used for computing taxonomic similarity, relational-based similarity and relatedness. The taxonomic similarity and relational-based similarity have the same algorithm complexity since they are both dependent on the computational complexity of the concept's IC value. Although the IC function varies from one baseline to another, yet their complexity is linear, that is $\mathcal{O}(n)$, where $n$ is the number of hierarchical features used to compute the IC value for each baseline (e.g. hyponyms, siblings, direct hyponyms, depth of concepts in the graph, etc.) as shown in  Table \ref{tb:icMitrics}. However, the complexity of the relatedness measure is $O(nm)$, where $n$ and $m$ represent the number of paths and the maximum path length between two terms respectively. Therefore, the overall complexity of the proposed algorithm is given by $O(mn)$, which is considered a reasonable polynomial complexity compared to the state of the art Knowledge Graph Embedding semantic similarity methods, where models need to be trained on large datasets using computationally expensive neural network configurations.

\section{Conclusion} \label{sec:Conclusion}
In this article, we examined the concept of semantic similarity based on the information content of terms. We introduced a novel approach that can be applied to any knowledge domain. The proposed approach exploits both taxonomic and non-taxonomic relations to compute IC and SemIC of all terms. These are employed at different granularity levels to measure semantic similarity. Furthermore, we introduced a new approach to measure relatedness based on weighted paths built out of non-taxonomic relations. 

The experimental results prove that non-taxonomic relations add valuable information to their associated terms, and contribute to determining the semantic similarity between them. Furthermore, it was shown that prevalence of each relation type is an important ingredient in measuring semantic similarity and relatedness, thus mimicking human perception. Therefore, we can conclude that non-taxonomic relations play a vital role in determining domain specific semantic similarity. 

The proposed technique can be applied to many research domains such as Information Retrieval (IR), Semantic Recommender Systems, and Natural Language Processing (NLP). For example, in the social media, non-taxonomic relations are dominant, and can be used to infer new insights about entities in the semantic graph (i.e., friends, places, products, and services). However, the proposed approach has a limitation, in the sense that it cannot be applied to a knowledge graph that has a few or no non-taxonomic relations. 


\bibliography{mycollection}

\pagebreak
\section*{Biography}
\InsertBoxL{0}{\includegraphics[width=1in,height=1.25in,clip,keepaspectratio]{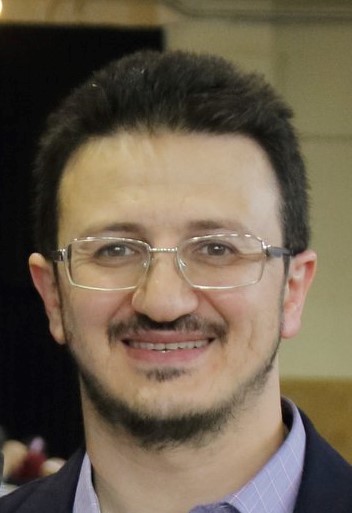}}[-1]
\textbf{Mohannad AlMousa} received his Bachelor Degree from Ryerson University (Canada) in the field of Information Technology Management (specialized in Knowledge and database) in 2010. He was then awarded a Master's Degree in Computer Science from Lakehead University (Canada) in 2014. Currently he is a PhD candidate in Software Engineering at Lakehead University. Mr. AlMousa's current dissertation is focused on video segmentation based on semantic similarity. His main research interests include Semantic knowledge representation, Recommender Systems, Natural Language Processing, and Knowledge Graphs.
\\
\InsertBoxL{0}{\includegraphics[width=1in,height=1.25in,clip,keepaspectratio]{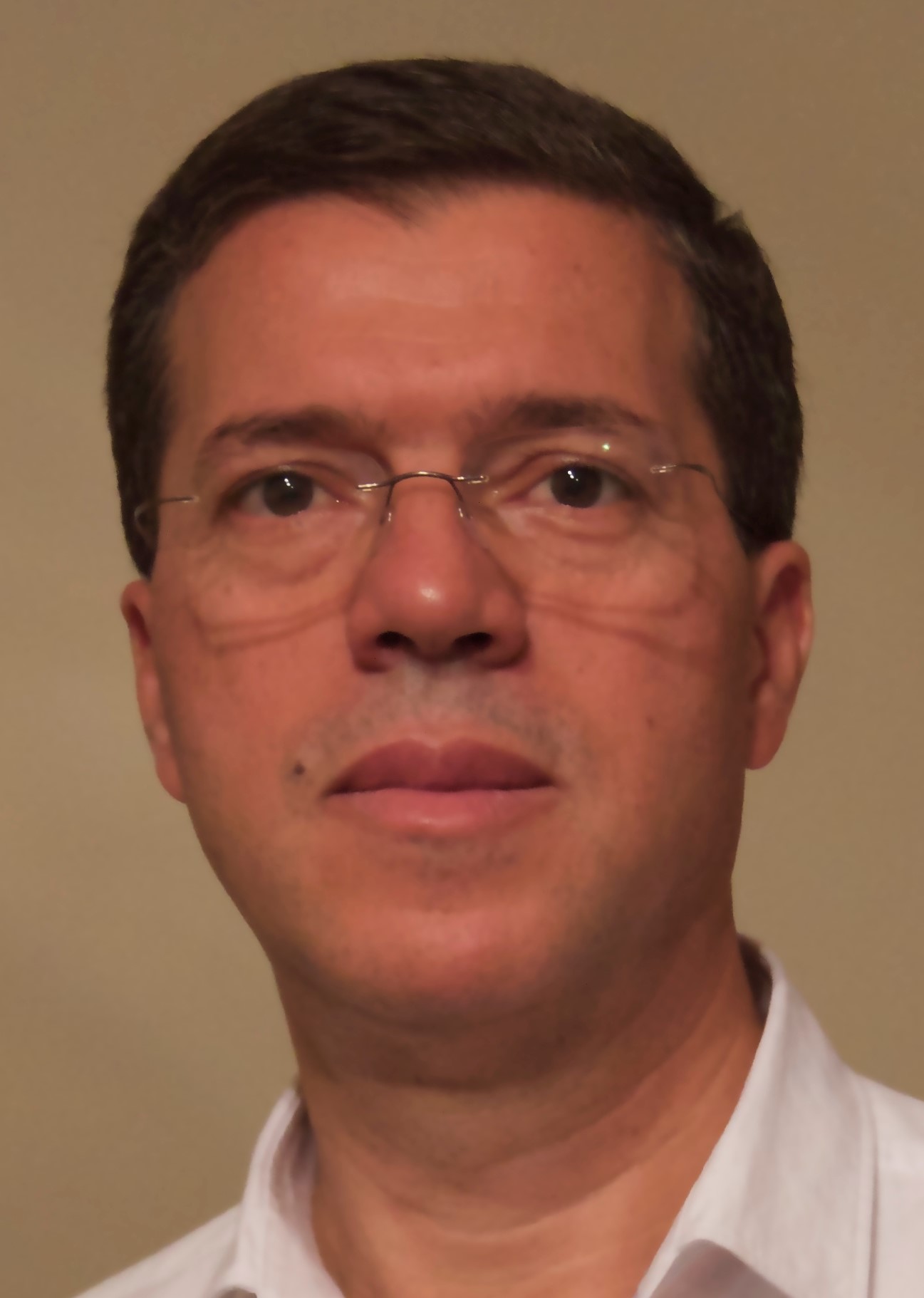}}[-1] \textbf{Rachid Benlamri} is a Professor of Software Engineering at Lakehead University - Canada. He received his Master's degree and a PhD in Computer Science from the University of Manchester - UK in 1987 and 1990 respectively. He is the head of the Artificial Intelligence and Data Science Lab at Lakehead University. He supervised over 80 students and postdoctoral fellows. He served as keynote speaker and general chair for many international conferences. Professor Benlamri is a member of the editorial board for many referred international journals. His research interests are in the areas of Artificial Intelligence,Semantic Web,  Data Science, Ubiquitous Computing and Mobile Knowledge Management.
\\
\InsertBoxL{0}{\includegraphics[width=1in,height=1.25in,clip,keepaspectratio]{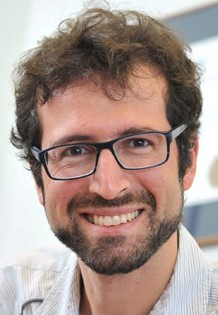}}[-1] \textbf{Richard Khoury} received his Bachelor’s Degree and his Master’s Degree in Electrical and Computer Engineering from Laval University (Québec City, QC) in 2002 and 2004 respectively, and his Doctorate in Electrical and Computer Engineering from the University of Waterloo (Waterloo, ON) in 2007. From 2008 to 2016, he worked as a faculty member in the Department of Software Engineering at Lakehead University. In 2016, he moved to Université Laval as an associate professor. Dr. Khoury’s primary areas of research are data mining and natural language processing, and additional interests include knowledge management, machine learning, and artificial intelligence.

\end{document}